\documentclass[sigconf]{acmart}

\def\BibTeX{{\rm B\kern-.05em{\sc i\kern-.025em b}\kern-.08emT\kern-.1667em\lower.7ex\hbox{E}\kern-.125emX}}
      
\copyrightyear{2021}
\acmYear{2021}
\setcopyright{acmlicensed}
\usepackage{verbatim}  
\usepackage{subfigure}
\usepackage{multirow}
\usepackage{threeparttable}

\usepackage{microtype}
\usepackage{graphicx}
\usepackage{booktabs} 

\usepackage{hyperref}

\usepackage{algorithm}  
\usepackage{algorithmicx}  
\usepackage{algpseudocode}  
\usepackage{bm}

\usepackage{bm} 
\usepackage{url}  
\usepackage{amsmath}
\newtheorem{theorem}{Theorem}[section]
\newtheorem{lemma}{Lemma}[section]
\newtheorem{corollary}{Corollary}[section]
\newtheorem{defn}{Definition}[section]
\algblock{ParFor}{EndParFor}
\usepackage{colortbl}
\algnewcommand\algorithmicparfor{\textbf{parfor}}
\algnewcommand\algorithmicpardo{\textbf{do}}
\algrenewtext{ParFor}[1]{\algorithmicparfor\ #1\ \algorithmicpardo}
\algrenewtext{EndParFor}{\textbf{end parfor}}

\begin{document}

%
\title{An Accurate and Efficient Large-scale Regression Method through Best Friend Clustering}

\author{Kun Li}
\affiliation{%
  \institution{Institute of Computing Technology,\\ Chinese Academy of Sciences} 
  \institution{University of Chinese Academy of Sciences} 
  \city{Beijing}
  \state{China}}
\email{likungw@gmail.com}
\author{Liang Yuan}
\affiliation{%
  \institution{Institute of Computing Technology,\\ Chinese Academy of Sciences} 
  \city{Beijing}
  \state{China}}
\email{yuanliang@ict.ac.cn}
\author{Yunquan Zhang}
\affiliation{%
  \institution{Institute of Computing Technology, \\Chinese Academy of Sciences} 
  \city{Beijing}
  \state{China}}
\email{zyq@ict.ac.cn} 
\author{Gongwei Chen}
\affiliation{%
  \institution{Institute of Computing Technology, \\Chinese Academy of Sciences} 
  \institution{University of Chinese Academy of Sciences} 
  \city{Beijing}
  \state{China}}
\email{gongwei.chen@vipl.ict.ac.cn} 
 

\begin{abstract}

As the data size in Machine Learning fields grows exponentially, it is inevitable to accelerate the computation by utilizing the ever-growing large number of available cores provided by high-performance computing hardware. However, existing parallel methods for clustering or regression often suffer from problems of low accuracy, slow convergence, and complex hyperparameter-tuning. Furthermore, the parallel efficiency is usually difficult to improve while striking a balance between preserving model properties and partitioning computing workloads on distributed systems. In this paper, we propose a novel and simple data structure capturing the most important information among data samples. It has several advantageous properties supporting a hierarchical clustering strategy that is irrelevant to the hardware parallelism, well-defined metrics for determining optimal clustering, balanced partition for maintaining the compactness property, and efficient parallelization for accelerating computation phases. Then we combine the clustering with regression techniques as a parallel library and utilize a hybrid structure of data and model parallelism to make predictions. Experiments illustrate that our library obtains remarkable performance on convergence, accuracy, and scalability. 

\end{abstract}

%
%

%

\maketitle
\section{\label{intro}Introduction}
Machine Learning (ML) has become one of the crucial mainstays of information technology over past decades, albeit commonly hidden, part of modern life pervasively. 
As technologies based on machine learning like artificial intelligence (AI) rise prominently, the data size that researchers have to study with is also growing exponentially~\cite{chen2020scene,li2019openkmc}. This makes training time for models range from hours to week, which poses intense pressures across computation, networking, and storage. 
Today, High Performance Computing (HPC) and parallel optimization techniques catalyze the modern revolution in machine learning. More and more companies are turning to HPC as the solution for ML-based productivity and  AI-enabled innovation, such as Google Cloud TPU, Amazon AWS AIHPC, and Microsoft Azure. Within the field of machine learning, clustering and regression are two fundamental and crucial techniques, which are distinguished as the representative unsupervised and supervised learning methods~\cite{draper1998applied,kuhn2015caret,ding2004k}.

Clustering methods divide data into different groups by data attributes. The aim is that samples in the same group are similar to each other and different from those in other groups. 
Basically, various types of clusterings are distinguished as hierarchical (nested) versus partitional (unnested) methods, where Agglomerative Hierarchical Clustering (AHC) and K-Means are two typical methods that are widely used. 
AHC and its variants make a set of nested clusters organized as a hierarchical tree~\cite{guha1998cure,day1984efficient,murtagh2012algorithms}. K-Means is simple and efficient on a variety of problems~\cite{dhillon2004kernel,wagstaff2001constrained,likas2003global,kanungo2002efficient}. 
Recently, Graph Clustering (GC) such as Minimum Spanning Tree clustering~(MSTC) also stands prominently for detecting clusters with irregular boundaries by graph theory. It uses either the Prim’s algorithm or Kruskal’s algorithm to construct a minimum spanning tree~(MST) first, sorts edges to remove inconsistent ones for connected components, and repeats until a threshold loops~\cite{kleinberg2006algorithm,grygorash2006minimum}. 

Regression involves intensive computations in the model training phase inherently. A standard distributed method for regression is to map the computation on $p$-processor system evenly and then perform a global reduction regularly~\cite{chu2007map,teo2010bundle,sun2012study}. However, these distributed regression methods work poorly in scaling cases due to busy synchronization sweeps. Well-known approaches use Divide-and-Conquer (DC) algorithm to optimize the parallel process, such as DCKRR and DCSVM~\cite{hsieh2014divide,zhang2013divide}. The basic idea of Divide-and-Conquer Regression (DCR) is to divide the data into $p$ similar parts, generate $p$ similar training models, and average $p$ models for a final solution. The memory and computation overhead is reduced by DCR, while the accuracy is not guaranteed in various cases~\cite{you2018accurate}.

More recent work is the parallel optimization for KRR: Balanced KRR v2 (BKRR2), which is the latest version by researchers~\cite{you2020fast}. BKRR utilizes K-Means to partition $n$ samples to $p$ clusters, where $p$ is the number of processes. Each cluster contains $n/p$ dispatched samples, and then $p$ models are generated correspondingly. The regression results on distributed models are gathered for an average. 
Based on the BKRR, BKRR2 is then proposed to improve the accuracy by training models independently without a global reduction.
Although BKRR2 proves that it could achieve higher accuracy and efficiency than the current fastest method~\cite{you2018accurate,you2020fast}, various problems are still not clearly addressed. 

First, the accuracy and the convergence for K-Means are not always
satisfactory on the large-scale dataset. It requires clustering iteration at least thousands of times for the experiments in BKRR2. Since the clustering center is changing dynamically, the volume of transferred data is massive in each iteration of the parallel clustering process for K-Means. BKRR2 also avoids confronting this problem directly and implement K-Means only with a process. Second, since the K is a hyperparameter to assign \#clusters in K-Means, its value is simply set to \#processes  in BKRR2 and that is not robust. The clustering results deviate from the real distribution drastically as \#processes change, which leads a poor accuracy in many cases. Even worse, the training time increases with more generated clusters assigned by increasing processes, which is essentially not adequate for large-scale training. Furthermore, BKRR2 achieves load-balance at the cost of accuracy. In the process of clustering, samples are traversed and dispatched to their nearest cluster center. However, a more relevant sample to some "full" cluster cannot be added to it for its late traversal order by the design of BKRR2. 

In this paper, we design an efficient parallel regression library to address the pending problems in existing methods. 
First, we propose a new graph structure called Best Friend Graph to capture the most important information among the data samples.  Based on the proposed structure, a novel clustering method, Best Friend Clustering (BFC), is presented. It reduces computation over AHC approaches, improves accuracy to K-Means methods, and enhances scalability than MST-based algorithms. 
Then we dig out the inherent properties of the graph structure and design a well-defined metric to determine the optimal aggregation.

Next, a balanced partition algorithm for data parallelism is proposed  based on the idea of backtracking. The partition results reflect the real distribution of clusters to the fullest extent. Instead of deleting edges by a non-increasing order serially in MST-based work~\cite{bateni2017affinity,grygorash2006minimum}, we gather samples in the same group by swapping their pointers simply. The larger clusters are then spilt by backtracking and smaller clusters are merged from the optimal hierarchy to obtain total $n/p$ samples in each process. Thus, the storage for edges is released, and the procedure could be parallelized easily.

Furthermore, Best Friend Clustering is applied to parallel regression. Although the above-proposed methods can be used separately, we combine them with multiple regression techniques as a complete parallel regression library. Each process generates  single or multiple training models independently according to the clustering results. For each test sample, it selects the best one from these models straightforward in the prediction phase. Thus, model parallelism is achieved efficiently in each process. To be as general as possible, our library has no specific limits on the dimensionality of the dataset and the format of the distance measure.

At last, we present an experimental study on our Best Friend Clustering with three different clustering methods. Then the accuracy and scalability are analyzed for our regression library with state-of-the-art work. The encouraging performance demonstrates the distinct superiority of the proposed methods on distributed systems.


The main contributions in this paper include:
\begin{itemize}
\item A novel Best Friend Clustering method is proposed for large-scale clustering, which is accurate, fast, and parameter-free.
\item Theoretical properties of the proposed clustering method are studied thoroughly, and we design a strategy to decide the optimal clustering aggregation through a well-defined metric based on analytical properties. 
\item We optimize the clustering on distributed systems, apply it to parallel regression, and propose a partition algorithm for load-balancing and spatial locality.
\item By utilizing a hybrid structure of data and model parallelism, we combine the proposed methods with regression techniques as a parallel library. It shows a remarkable performance on convergence, accuracy, and scalability.

\end{itemize}

\section{Background}
\subsection{Regression Techniques}
\label{bg:rg}
Regression is a supervised machine learning technique that is utilized to investigate the relationship between one or more predictors and response variables for a best-fit curve.
Basically, there are various algorithms that are utilized to build a regression model, and in our paper, three extensively-used techniques are molded out and implemented in our library.
\subsubsection{Linear regression}
Linear regression is a widely-used method to find the linear relationship between the dependent variable and one or more independent variables by employing a straight line. Given a d-dimensional training sample $\textbf{x}_i$ and the corresponding measured regressand $y_i$, the objective in training phase is to find a $\textbf{w}$ such that $y_i\approx \textbf{w}^T\cdot\textbf{x}_i$. This can be formulated as a least squares problem in Equation \ref{lstsq} to find the optimal line by minimizing the sum of the residuals. Then for a test sample $x^*$, we can utilize the training model to predict the regressand by $y^*=\textbf{w}^T\cdot \textbf{x}^*$, and evaluate the accuracy with Mean Squared Error (MSE) method.
\begin{equation}
\label{lstsq}
    \textbf{w}_{opt}=\arg \min \frac{1}{2}{\sum}_{i}^n(y_i-\textbf{w}^T\textbf{x}_i)^2
\end{equation}

\subsubsection{Kernel Ridge regression}
To avoid overfitting and solve some ill-posed problems, L2 regularization with a positive parameter $\lambda$ is used in Equation \ref{ridgerg}, which is called ridge regression.
\begin{equation}
\label{ridgerg}
    \textbf{w}_{opt}=\arg \min \frac{1}{2}{\sum}_{i}^n(y_i-\textbf{w}^T\textbf{x}_i)^2+\frac{1}{2}\lambda||\textbf{w}||^2
\end{equation}
Moreover, kernel method is widely utilized to map samples to a high dimensional space using a nonlinear mapping. Thus the Kernel Ridge Regression (KRR) is presented by combining ridge regression with kernel method, and it learns model in high dimensional space for a better prediction accuracy. The solution vector \bm{$\alpha$} can be written in closed form in Equation \ref{eq:krr}, 
\begin{equation}
\label{eq:krr}
    \bm{\alpha}=(\textbf{K}+\lambda \textbf{I}_N)^{-1}\textbf{y}
\end{equation}
where \textbf{K} is a n-by-n kernel matrix constructed by \textbf{K}$_{i,j}=\phi(x_i,x_j)$, $\textbf{y}$ is the corresponding n-by-1 regressand vector. Then \bm{$\alpha$} is employed to predict regressands in prediction phase as Equation \ref{eq:krrpd}.
\begin{equation}
\label{eq:krrpd}
     y^*={\sum}^N_{i=1}{\alpha}_ik(x^*,x_i)
\end{equation}

\subsubsection{Support Vector Regression}
Support Vector Machine (SVM) can also be used as a regression method called Support Vector Regression (SVR), holding all the key features such as maximal margin that characterize the algorithm.
Both KRR and SVR can learn a non-linear model by using kernel tricks, while they differ in the loss functions, i.e., ridge and epsilon-insensitive loss respectively. In the case of SVR, a margin of tolerance (epsilon) is set to the SVM, and we can tune it to gain the desired accuracy of our model. 
Moreover, slack variables $\xi$ and $\xi^*$ can also be added to guard against outliers like SVM. The solution is obtained in Equation \ref{eq:svrslacks} that is subject to Equation \ref{eq:svrslackcons}, where \textbf{w} is the magnitude of the normal vector to the surface that is being approximated, C is a tuneable regularization.
\begin{equation}
\label{eq:svrslacks}
   \textbf{w}_{opt}=\arg \min \frac{1}{2}||\textbf{w}||^2+C{\sum}_{i=1}^{N}\xi_i+\xi_i^*
\end{equation}
\begin{equation}
\begin{aligned}
\label{eq:svrslackcons} 
 y_i-\textbf{w}^Tx_i\leq\varepsilon+\xi_i^*\\
   \textbf{w}^Tx_i-y_i\leq\varepsilon+\xi_i\\
   \xi_i,\xi_i^*\geq 0     
\end{aligned}
\end{equation}

\subsection{\label{bg:cm}Clustering Methods}
Many clustering methods are proposed in the literature to recognize the cluster of different characteristics. In this subsection, AHC, K-Means, and MSTC are introduced briefly.

\subsubsection{Agglomerative Hierarchical Clustering}

Algorithm \ref{alg:ahc} describes AHC formally. Generally AHC algorithm can produce a better-quality clustering result. However, the slow convergence speed, extensive computation operations, and expensive storage requirements make scaling problematic on a larger dataset. Moreover, since no global objective function is directly minimized, it remains to be determined when to stop merging, and the clustered decisions cannot be undone.
\begin{algorithm}[htb]
    \caption{Agglomerative Hierarchical Clustering Algorithm}
    \label{alg:ahc}
    \begin{algorithmic}[1]  
        \State Let each sample be a cluster.
        \Repeat
            \State Merge the nearest two clusters as a new one.
            \State Update the proximity matrix.
        \Until{Only one cluster remains.}
    \end{algorithmic}
\end{algorithm}

\subsubsection{K-Means}

The basic steps are illustrated in Algorithm \ref{alg:kmeans}. K-Means is simple and efficient on a variety of problems. However, non-globular clusters or clusters of different  densities and sizes cannot be solved by K-Means, and the outliers in data can also affect results. Furthermore, $K$ value is a hyperparameter specified in advance, which determines the quality of clustering significantly.
\begin{algorithm}[htb]
    \caption{K-Means Algorithm}
    \label{alg:kmeans}
    \begin{algorithmic}[1]  
        \State Choose K initial sample as centroids.
        \Repeat
            \State Assign each sample to its nearest centroid to form K clusters.
            \State Recompute K centroids.
        \Until{Centroids remains unchange.}
    \end{algorithmic}
\end{algorithm}
\subsubsection{MST-based Clustering}

The basic idea of MSTC is as follows. 
The algorithm can obtain a comparatively better result on clusters with irregular boundaries, while it is highly sequential and computationally intensive~\cite{bateni2017affinity}.
 
\begin{algorithm}[htb]
    \caption{MST-based Clustering Algorithm}
    \label{alg:mstc}
    \begin{algorithmic}[1]  
        \State Compute and sort the pairwise distances.
        \Repeat
            \State Run MST algorithm to form clusters. 
        \Until{$k$ clusters.} 
    \end{algorithmic}
\end{algorithm}

\section{Method}
In this section, we first propose a simple, efficient, and accurate
method for large-scale clustering. Section \ref{coupleclustering} and Section~\ref{prop} provide the definition of Best Friend Graph and properties of this new method. They serve as the fundamental data structure of our method. Then a strategy to decide the optimal clustering aggregation through a well-defined metric is presented in Section~\ref{method-2}. Next, we use a hybrid structure of data and model parallelism for parallel regression. Section \ref{method-3} presents the detailed description of balanced partition for data parallelism. It achieves load-balance by utilizing merge and split operations on the hierarchical clustering structure. At last, the parallel regression method achieved with model parallelism is described in Section \ref{method-4} .






\subsection{Best Friend Clustering}
\label{coupleclustering}
Clustering is intuitively inspired by the most fundamental social relation, friend circle, for detecting the potential groupings in the data. Existing clustering methods normally consider relationships between all pairs of samples. This leads to a slower convergence rate and a large overhead, which are not adequate for large-scale datasets on distributed systems.
In reality, many of the relationships can be ignored. Our clustering method arises from the observation that
people often make a decision with their best friends. Therefore, we simplify the friend cycle by only considering the most important best friend relationship. 

\begin{figure}[ht]
\begin{center}
\centerline{\includegraphics[width=\columnwidth]{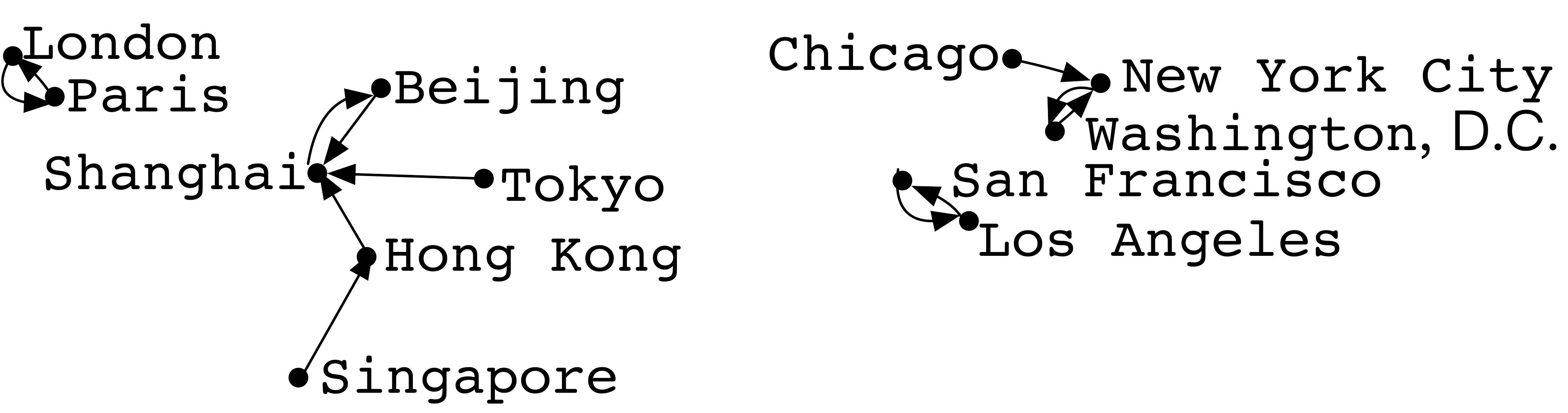}}
\caption{Schematic diagram for Best Friend Graph with a case study of global cities.}
\label{cities_s1_directed}
\end{center}
\end{figure}

A \emph{Best Friend Graph} $G(V,E)$ is defined on a dataset of
$n$ input samples for clustering.
The vertex set $V$ consists of  these $n$ data elements.
Each vertex $i\in V$ is associated with a 
directed \emph{best friend edge} $(i,j)$ where the destination $j$ is its best friend, i.e. $i$'s nearest neighbor.
Note that if there exist multiple nearest neighbors with the same distance to $i$, we simply
choose the vertex with the smallest lexicographic order as the only best friend of $i$.
Therefore, a Best Friend Graph $G(V,E)$
with $|V|=n$ vertices contains $n$ directed best friend edges.
Figure \ref{cities_s1_directed} illustrates a tractable example
of the Best Friend Graph. It contains top 12 city data by Global Cities Index~\cite{c.elmohamed} 
and 12 directed edges identifying the best friend relationships among them.

\begin{figure}[htb]
\begin{center}
\centerline{\includegraphics[width=\columnwidth]{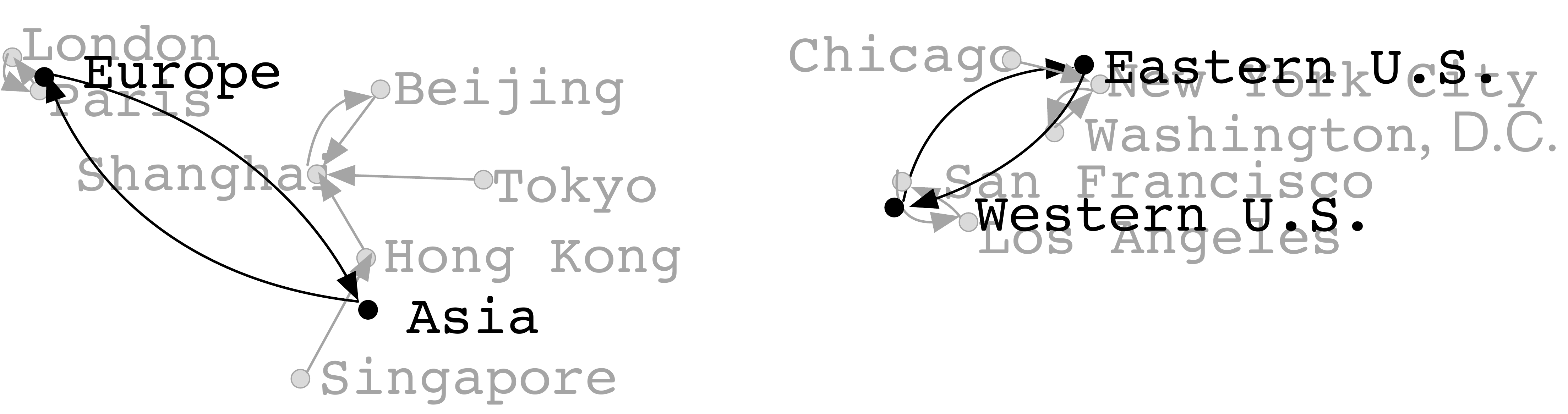}}
\caption{Schematic diagram for Best Friend Clustering with a case study of global cities.}
\label{cities_s2_directed}
\end{center}
\vskip -0.2in
\end{figure}

Based on the Best Friend Graph, a cluster is defined as a group of connected vertices. 
Evidently, the city samples are clustered into four groups by related friends in the first step.
The clusters distinguished by Best Friend Clustering are consistent with the geographic taxonomies. For example, Los Angeles and San Francisco are recognized as the Western US. Chicago, Washington DC, and New York City are then classified as the Eastern US. 
Specifically, each new cluster is represented by a center point sample that is computed as the mean value of all samples in it.
Let $T_k(V_k,E_k)$ be a connected subgraph $k$ in $G$ and $v_k=|V_k|$. The center point $\overline{x}_k$ representing $T_k$
is defined  by Equation~\ref{eq:newcc}.

\begin{equation}
\label{eq:newcc}
    \overline{x}_k=\frac{1}{v_k}\sum_{x_i\in V_k}x_i 
\end{equation}

We can then build a new Best Friend Graph of all generated cluster centers in the previous step.
In our example, the four clusters are combined into two groups as illustrated in Figure \ref{cities_s2_directed}.
We recursively apply this approach to the new clusters  until there 
is only one cluster containing all the samples. 
The whole process converges quickly in logarithmic time as a 
cluster contains at least two samples in each hierarchy. 
Furthermore, the computation and communication time are 
reduced progressively with hierarchies.

\subsection{\label{prop}Properties}
In this subsection, we analyze several crucial properties of Best Friend Graph, which are the basis to decide the optimal aggregation.

\begin{lemma}
\label{lemma-1}
There exists at least one directed cycle in a Best Friend Graph.
\end{lemma} 

\emph{Proof.} If we replace all directed edges
with undirected edges, a Best Friend Graph becomes an undirected graph with $n$ vertices and $n$ edges.
By induction, we can easily prove that it involves at least one undirected cycle.
Transforming the edges on the cycle back to
directed edges, we obtain a directed cycle.
Otherwise, there must be one vertex $i$ associated
with two best friend edges $(i,*)$
on the cycle, a contradiction to the best friend
graph definition.

\begin{lemma}
\label{lemma-2}
Each weakly connected component of a Best Friend Graph contains one and only one directed cycle.
\end{lemma}

\emph{Proof.} Each weakly connected component
$G'(V',E')$
of a Best Friend Graph $G(V,E)$ is still a Best Friend Graph  according to the definition. 
With Lemma \ref{lemma-1}, $G'$ contains at least one cycle.
Thus we obtain $|E'|\geq |V'|$.
We must have $|E'| = |V'|$, i.e. only one cycle
in $G'$,
otherwise will get $|E| > |V|=n$, contradicting
the Best Friend Graph definition.

\begin{lemma}
The edge weights on a directed path
of a Best Friend Graph
are non-increasing.
\label{lemma-non-increasing}
\end{lemma} 

This is obvious from the definition.
For example, for the directed path $i\rightarrow j \rightarrow k$, we have 
$\omega(i,j) \geq \omega(j,k)$,
otherwise $j$'s best friend should be $i$ rather than $k$.

\begin{lemma}
\label{chain}
\label{lemma-3}
The length of a directed cycle in
a Best Friend Graph is two
and the weights of edges on a cycle
are identical.
\end{lemma} 

\emph{Proof.} Assume there is only one nearest neighbor to each vertex,
i.e. the weight $\omega$ of a best friend edge $(i,j)$ is strictly smaller than the distance
between $i$ and other vertex $\omega(i,j) < \omega(i,k), k\neq i,j$.
If there exists a directed cycle involving more than two 
vertices, e.g. $i\rightarrow j \rightarrow k\rightarrow i$, we have 
$\omega(i,j) > \omega(j,k) > \omega(k,i)=\omega(i,k)$.
Thus the best friend edge of $i$ should be $(i,k)$, a contradiction.

Otherwise, we have $\omega(i,j) \geq \omega(j,k)\geq \omega(k,i)=\omega(i,k)$.
Since $(i,j)$ is a best friend edge,
we obtain $\omega(i,j)=\omega(i,k)$ 
and $\omega(i,j) = \omega(j,k)= \omega(k,i)$.
According to the smallest lexicographic order
rule in Best Friend Graph definition,
we get an inconsistent lexicographic order
$j\prec k\prec i \prec j$.

\begin{corollary}
\label{corollary-1}
Starting from any vertex and traversing along 
the Best Friend Graph will enter the cycle.
\end{corollary}

\begin{defn}
\label{def1}
A best friend forest $F(V,E)$ of a Best Friend Graph $G(V,E')$ is defined by replacing all directed edges with undirected ones and removing one edge on each cycle. 
\end{defn}

\begin{theorem}
\label{gspt}
A connected component, a tree $T(V_T,E_T)$ in a best friend forest is a 
minimum spanning tree of the corresponding complete graph
$G(V_T, E)$. 
\end{theorem} 

\emph{Proof.} Let $T^*(V_{T},E_{T^*})$ be the MST of 
the complete graph
$G(V_T, E)$. For an edge $(i,j)$ in $E_{T^*}-E_{T}$,
we obtain an undirected path in $T(V_T,E_T)$.
According to Corollary~\ref{corollary-1},
the corresponding directed path connecting $i$ and $j$ 
in the Best Friend Graph is one of the following three cases:
 $i\to\dots\to j$,
  $i\leftarrow\dots\leftarrow j$
  and $i\to\dots\to k \leftarrow\dots\leftarrow j$.
Since adding $(i,j)$ it to $T(V_T,E_T)$ leads to a cycle,
there must be an edge $(u,v)$ in the path that does not
belong to $T^*(V_{T},E_{T^*})$.
In any of the three path cases, we have $\omega(i,j)\geq\omega(u,v)$
according to Lemma \ref{lemma-non-increasing}. 
We must have $\omega(i,j)=\omega(u,v)$,
otherwise we will get spanning tree with a smaller weight by
replacing $(i,j)$ with $(u,v)$.
This contradicts the assumption that $T^*(V_{T},E_{T^*})$ is a MST.
Repeat the process for all edges in $(i,j)$ in $E_{T^*}-E_{T}$,
we will obtain $T$ with a same weight to $T^*$.
Thus $T(V_{T},E_{T})$ is also a MST.

\begin{figure}[!t]
\begin{center}
\centerline{\includegraphics[width=\columnwidth]{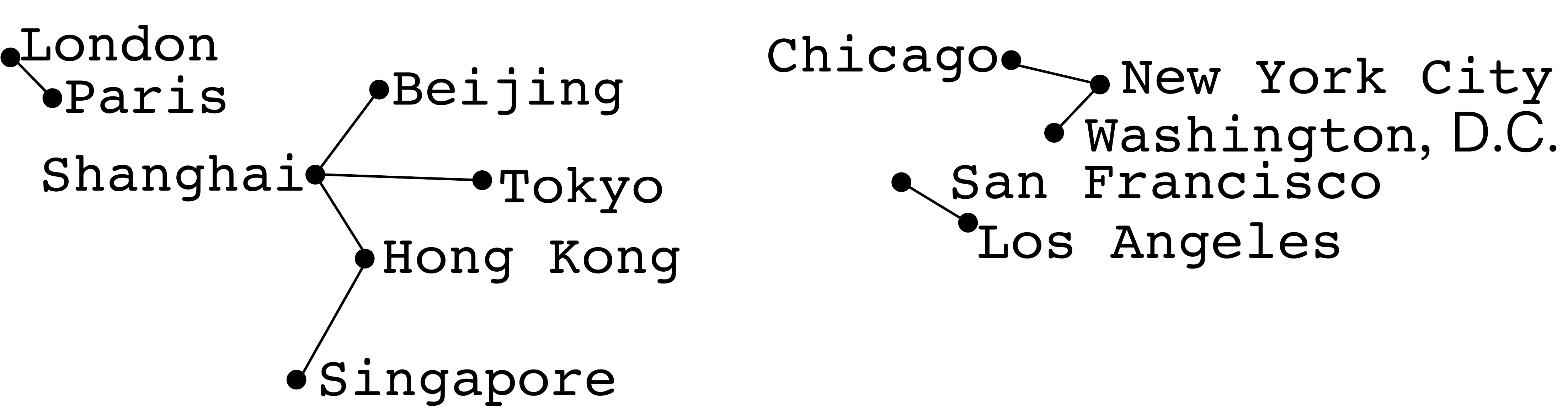}}
\caption{Best friend forest with a case study of global cities.}
\label{cities_s1_undirected}
\end{center}
\vskip -0.2in
\end{figure}


Figure \ref{cities_s1_undirected} shows the 
corresponding best friend forest of the example in Figure \ref{cities_s1_directed}.
With Theorem~\ref{gspt}, each connected component in Figure \ref{cities_s1_undirected} is a minimum spanning tree, and the graph 
is a minimum spanning forest $F(V,E)$.

\subsection{Optimal Aggregation}
\label{method-2}



The target of clustering is expected to make high intracluster compactness and intercluster dispersion~\cite{asano1988clustering}.
Since a set of clustering hierarchies are built,
we turn to choose an optimal clustering level as the input to our regression model by using a rational metric.
As discussed in Theorem~\ref{gspt}, Best Friend Clustering specifies a minimum spanning forest for each hierarchy intrinsically, and it connects all scattered clusters as a whole network.
Thus, the clustering validity is analyzed based on the MST network, where metrics are quantified by the properties of MST.

\begin{defn}
\label{defcompact}
Let a $T_i=(V_i,E_i)$ denote a minimum spanning tree in a best friend forest $F$. The intracluster compactness $c_i$ for $T_i$ is defined as:
\begin{equation}
    c_i=\frac{1}{e_i}\sum_{(j,k)\in E_i}\omega(j,k),
\end{equation}
 where $e_i=|E_i|$.
\end{defn}

\begin{algorithm}[!t]
    \caption{Calculating Compactness}
    \label{alg:friend}
    \begin{algorithmic}[1]  
         \Require  $visited[]=0,c[]=0,e[]=0,mst\_num=0$ 
                 \Function {InitializeBestFriendForest}{}
            \For{$i = 0 \to n-1$}
                \State $j = findBestFriend(i)$
   \State $addEdge(i, j, \omega(i,j))$
            \EndFor
        \EndFunction 
        \Function {TraverseMST}{}
            \For{$i = 0 \to n-1$}
                \If{$!visited[i]$}   
   \State \Call{Search}{$i$}
   \State $c[mst\_num]/=e[mst\_num]$
 \State $mst\_num = mst\_num+1$
                \EndIf
            \EndFor
        \EndFunction 
        \Function{Search}{$i$}
            \State  $visited[i]= 1$. 
            \State $e[mst\_num]+=1$
             \While{$getNextNeighbor(i,\&k)$}
                \If{($!visited[k]$)}   
            \State $c[mst\_num]+=getEdgeWeight(i,k)$
                \State \Call{Search}{$k$}
                \EndIf
                \EndWhile
        \EndFunction 
    \end{algorithmic}
\end{algorithm}

Algorithm \ref{alg:friend} shows the procedure of the best friend forest construction and the distance calculation of each connected component.
The first function finds all best friend edges and records them in a global array.
The second function
traverses all
connected components.
It finds an unvisited node as a root of a MST
and feeds it the  third function which 
utilizes the Depth-First Search (DFS)
to traverse all the nodes in the MST
and accumulate the number and their  weights
in two global arrays $c[]$ and $e[]$.


\begin{defn}
\label{defdisperse}
Let a best friend forest $F_k$ be the $k$th hierarchy produced by Best Friend Clustering. Assume that there are $m$ clusters {$T_1$, $T_2$, ..., $T_m$} in $F_k$, and cluster $T_i$ contains $v_i$ samples. Then the intercluster dispersion $d_i$ for $T_i$ is defined as:

\begin{equation}
\label{eq:defdisperse}
    d_i=\min\{d(\overline{x}_i,\overline{x}_j)|1\leq j\leq m,j\neq i\},
\end{equation}
\end{defn}
where $\overline{x}_i$ and $\overline{x}_j$ are the new cluster centers and $d(\overline{x}_i,\overline{x}_j)$ is the Euclidean distance between cluster $T_i$ and $T_j$.

The following metric combines the
intracluster compactness $c_i$ and the intercluster dispersion $d_i$.
It serves to determine the 
optimal clustering level.

\begin{defn}
\label{defhci}
Let MSTs {$T_1$, $T_2$, ..., $T_m$} denote clusters in a best friend forest $F_k$ for the $k$th dendrogram hierarchy dendrogram produced by Best Friend Clustering. Then the HCI(k) is defined as a linear combination of the intracluster compactness and intercluster dispersion:
\begin{equation}
    HCI(k)=\frac{1}{m}{\sum}_{i=1}^m(\frac{d_i-c_i}{d_i+c_i}),
\end{equation}
 and the optimal cluster number is:
\begin{equation}
    k_{opt}=\arg \max \{HCI(k)\}.
\end{equation}
\end{defn}


\begin{figure}[!t] 
\begin{center}
\centerline{\includegraphics[width=\columnwidth]{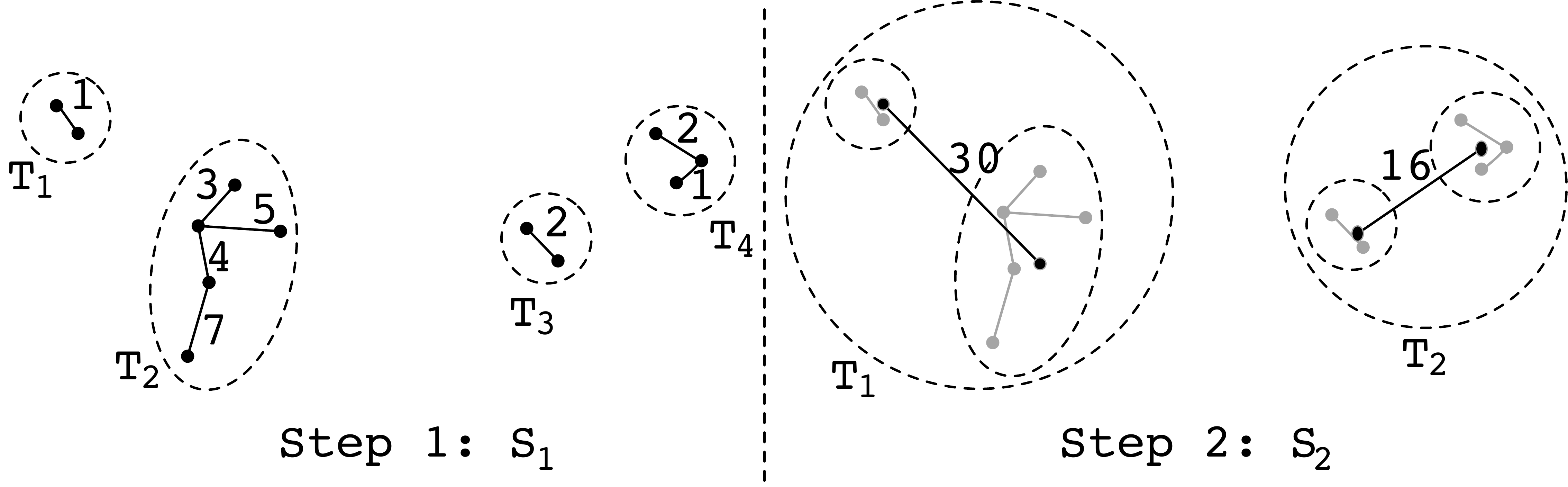}}
\caption{Hierarchical MST results by Best Friend Clustering with a case study of global cities.}
\label{wcities}
\end{center}
\vskip -0.3in
\end{figure}

With scaled distance denoted on each edge, hierarchical MST results are depicted in Figure \ref{wcities} for the first two clustering levels.
We have $c_1=1$, $c_2=(3+5+4+7)/4=4.75$, $c_3=2$, $c_4=(2+1)/2=1.5$ in $S_1$, and $c_1=30$, $c_2=16$ in $S_2$. Since similar vertices are merged iteratively, an important observation is that the $d_i$ for cluster $i$ in $S_k$ is exactly the best friend weight for vertex $i$ in $S_{k+1}$. For instance, $d_1=d_2=30$, $d_3=d_4=16$ are obtained in $S_1$ by the information acquired in $S_2$. Thus we have $HCI(1)=0.82$ and $HCI(2)=0.51$, and the clustering result of $S_1$ is evaluated better than $S_2$.

\subsection{Balanced Partitions}
\label{method-3}
\begin{figure*}[h]
\begin{center}
\centerline{\includegraphics[width=16cm]{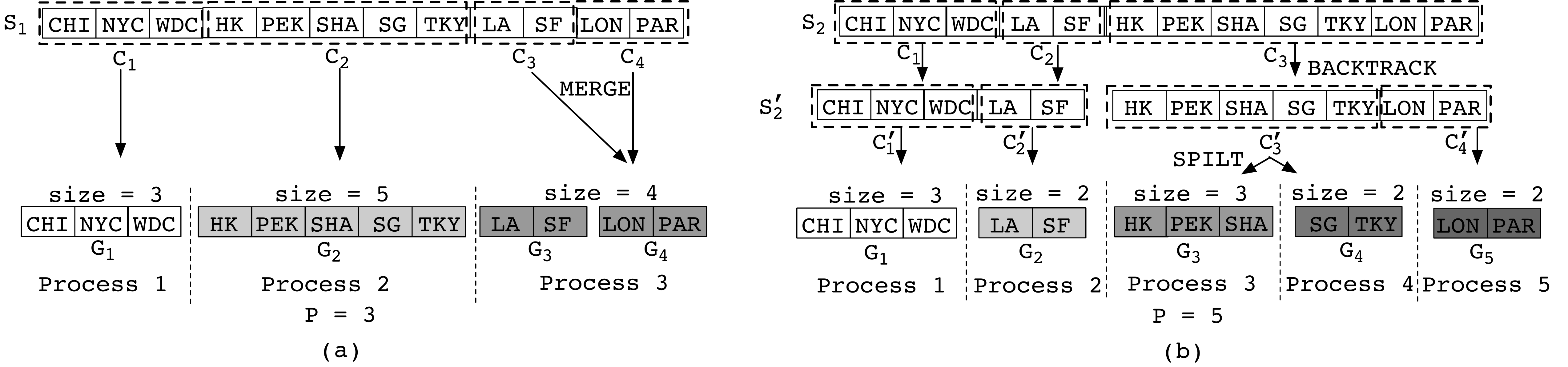}}
\caption{Balanced partitions to achieve load balancing on each process. Figure \ref{back} (a) describes the MERGE for small tasks while Figure \ref{back} (b) shows the SPLIT for large tasks based upon backtracking mechanism. Where appropriate, abbreviated forms are used for city names.}
\label{back}
\end{center}
\vskip -0.2in
\end{figure*}
Based on the profiling results, we observe that the partitions by clustering are typically irregular and imbalanced. This makes computing nodes load-imbalanced, and thus we need to devise a new partition algorithm to achieve data parallelism. In our design, a balanced partition on $p$ computing nodes means that the number of samples on each node is close to $n_p=n/p$. Based on the data organization of clustering result, we propose a balanced partition algorithm by utilizing a backtracking mechanism, which is composed of MERGE and SPLIT operations.

MERGE is performed on the piecemeal clusters with small sizes. Figure \ref{back} (a) shows a case with 4 processes on the clustering result $S_1$. Since we have $n_p=12/3=4$, the sizes of cluster $C_3$ and $C_4$ are too small compared to $n_p$. Therefore, we sort the $S_1$ by cluster size, and merge the small clusters into the same node for a total size close to $n_p$ on it. It is worth noting that the models are still trained independently on each node, which means the objective of MERGE is to make up the $n_p$ instead of mixing models on different clusters. In Figure \ref{back} (a), cluster $C_3$ and $C_4$ are merged into process 3 as group $G_3$ and $G_4$. The total sizes are 3, 5, and 4 respectively on process 1 to 3, which achieves a balanced partition for $S_1$.

\begin{algorithm}[!t]
    \caption{Parallel Regression}
    \label{alg:regression}
    \begin{algorithmic}[1]  
        \Require  {$n$ samples for training, $k$ samples for testing, best-clustered results $A$}
        \State $p\gets$ rank of a process
        \State Initialize best Mean Squared Error $MSE^*\gets\infty$ 
        \State Balanced Partitions on $n$ samples in $A$ for $A^p$
        \Function {Regression}{}
                \State \Call{Training}{$A^p$}
                \State \Call{Testing}{$M^p$}
                \State Reduce: MSE=$(\sum^P_{p=1}e^p)/k$
                \If{$(p=0)\&\&(MSE<MSE^*)$}    
  \State $MSE^*=MSE$
                \EndIf
        \EndFunction 
        \Function {Training}{$A^p$}
            \State $C^p\gets$ Cluster Number in $A^p$ on Rank $p$
            \For{$i = 0 \to C^p$}
                \State Training Model $M^p_i$ for Cluster $i$ on Rank $p$
            \EndFor
        \EndFunction 
        \Function {Testing}{$M^p$}
            \State Initialize local MSE $e^p\gets0$
            \For{$i = 0 \to k-1$}
                \If{FindNearCLuster$(i)=C^*$}   
  \State Select best $M^*$ for prediction 
  \State Update $e^p$
                \EndIf
            \EndFor
        \EndFunction 
    \end{algorithmic}
\end{algorithm} 

SPLIT is utilized to separate the large clusters of which the size is much larger than $n_p$. Since the sample pointers in same clusters are moved together in each iteration, we achieve SPLIT operation by backtracking mechanism based on the clustering array structure. A case with 5 processes on $S_2$ is illustrated in Figure \ref{back} (b). In this case, we have $n_p=12/5=2.4$ while the original cluster $C_3$ are larger than $n_p$ apparently. Thus we perform backtracking on the $S_2$ 
by the tracks of pointers. The backtracked $S_2^{'}$ contains 2 new split clusters, where the size of $C_3^{'}$ is still large than $n_p$. Then we perform another SPLIT to separate $C_3^{'}$ into $G_3$ and $G_4$ respectively. Therefore, the group $G_1$ to $G_5$ are dispatched to nodes evenly, and they are trained as independent models. By utilizing the backtracking mechanism, a principle is followed that closer samples are always guaranteed to gather together after the SPLIT operation, and spatial locality is also exploited simultaneously.

\subsection{Independent Prediction}
\label{method-4}
Parallel regression methods normally construct
$p$ independently models, where $p$ is the number of processors (hardware parallelism). 
They often take $p$ into consideration from the beginning.
Two major disadvantages exist with this approach.

First, $p$ is essentially irrelevant to the input data and  may mismatch the intrinsic structure of data samples.
Our method employs the Best Friend Graph hierarchically
and efficiently constructs a series of cluster levels
that does not depend on any predefined value.
Second, existing methods may require a data reorganization
to improve the load-balance.
However, this procedure may lose the relationship information among the data that is moved from one cluster to others
and hurts the compactness of the final models.

Unlike the existing work~\cite{zhang2013divide,you2020fast}, model parallelism is utilized in our work for making predictions independently on each process. For a given test sample $x^*$, we only use the corresponding model $M^*$ to perform a prediction if its closest cluster center is $C^*$ on process $p$. Instead of conducting communication regularly, the errors are accumulated on each node first, and only a Reduce operation is required at last to make statistical analyses. Since an intact message is cheaper than  scattered messages in MPI communication, the latency overhead is further reduced by this optimization. The parallel regression is summarized in Algorithm \ref{alg:regression} formally.

\section{Implementation}
\subsection{Parallelization}


To reduce the computation, memory, and communication overheads, we design an efficient parallel implementation of our method.
The parallelization of the first function of Algorithm \ref{alg:friend}
is straightforward.
The data samples are evenly distributed to all processors
and the calculation is parallelized accordingly.
 The distribution of the example is  shown in Figure \ref{parallel}
 where each process is dispatched with 3 samples for computing their nearest neighbors respectively.
 
 The traversal of the best friend forest
 and the calculation of compactness, i.e. the second function of 
  Algorithm \ref{alg:friend} seems to be an inherently serial task.
  However, provided with Lemmas in the previous section, we are able
   to identify individual components (trees) with the edge information
   in the best friend forest. With Lemma \ref{lemma-2}, we know that
   each cycle identifies a tree, and with Lemma \ref{lemma-3} a cycle is
   easy to find by searching its two equal edges.
   
   Therefore, all the best friend edges are gathered to all processors.
   Every processor then finds all pairs of equal edges by sorting
    all $|V|$ edges or using a hash method.
    The number of trees in the best friend forest is the number
    of pairs of equal edges. 
    Each processor 
    traverses a set of trees and 
    the traversal of each tree starts with either of the two nodes
    on its cycle.
   Overall, it only transfers short messages that only contain two values: the pairs of nearest neighbors and the corresponding shortest distances
  in our implementation. Algorithm \ref{alg:parallel} provides the parallel version
  of the second function TRAVERSEMST in Algorithm \ref{alg:friend}.

\begin{algorithm}[!t]
    \caption{Parallellization of TRAVERSEMST in Algorithm \ref{alg:friend}}
    \label{alg:parallel}
    \begin{algorithmic}[1]  
            \State Gather the whole Best Friend Forest
            \State $hash.initialize()$
            \State $mstNum=0$
            \For{$i = 0 \to n-1$}
            \State $j=getBestFriend(i)$
                \If{$hash.exist((i,j))$}   
   \State $startNode[mstNum] = i$
   \State $mstNum+=1$
                \Else
                \State $hash.add((i,j))$
                \EndIf
            \EndFor
            \ParFor{$i = 0 \to mstNum-1$}
            \State \Call{Search}{$startNode[i]$}
            \EndParFor
    \end{algorithmic}
\end{algorithm}




\subsection{Vectorization}
 The widely-used Euclidean distance is employed for distance calculation in  our work. 
 The process of finding the nearest neighbor dominates the overheads of the  computation.
 To further leverage the ability of vector processing units in modern CPUs, we group $vl$ data samples in a vector register and  perform $vl$ calculations in a SIMD style,
 where $vl$ is the vector length.
 This improves the computation efficiency significantly.


\subsection{Data Organization}

\begin{figure}[t]
\begin{center}
\centerline{\includegraphics[width=7cm]{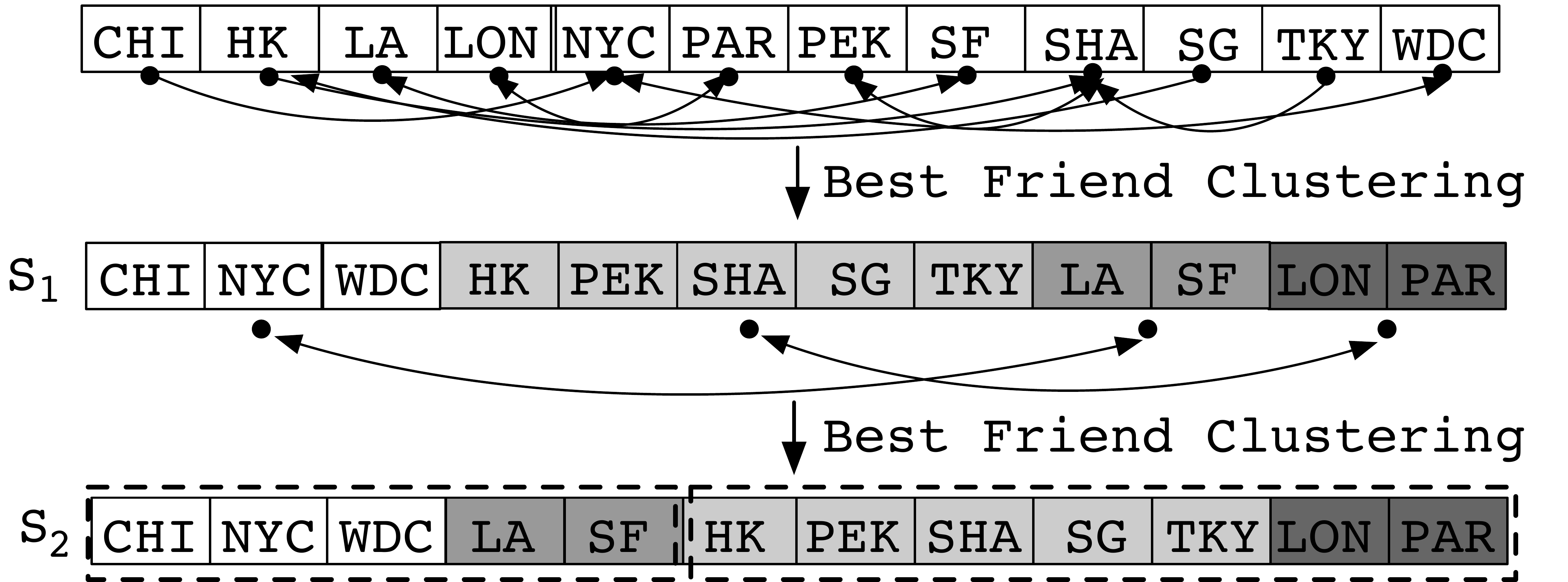}}
\caption{Data organization in Best Friend Clustering. Clusters are distinguished by colors and boxes in two steps respectively. }
\label{parallel}
\end{center}
\vskip -0.3in
\end{figure}

Provided with the load-balancing scheme described in 
Section \ref{method-3},
we design a simple data organization method
to guarantee the SPLIT operations produce
partitions with good compactness.
To this end,
sample pointers of the 
same cluster are swapped to stay together after calculating each clustering level. 
Except the first clustering level, each point actually represents a cluster.
The relative positions of points in a low-level cluster are fixed, i.e.
they are swapped as a whole big point.
This data organization is critical to balanced partition since it 
guarantees that the closer samples are in the array, 
the more similarities match. For example in Figure 
\ref{parallel}, although London, Paris, and Hong Kong are 
clustered into the same group in the second $S_2$ hierarchy, 
London and Paris contain more similarities as they are far 
away from Hong Kong in the clustering array.

\section{Experiments}

\subsection{Setup}

\begin{table}[htb] 
\caption{Description of Classic Shape Datasets.}
\label{shdataset}
\begin{center}\small 
\resizebox{\columnwidth}{!}{
\begin{sc}
\begin{tabular}{lcccr}
\toprule
Dataset & \#Samples & \#Labels & \#Dimensions  \\
\midrule
Zahn's Compound~\cite{zahn1971graph}   & 399&  6 & 2  \\
 Aggregation~\cite{gionis2007clustering}&788 &7&2 \\
 R15~\cite{veenman2002maximum}& 600& 15  &2\\ 
\bottomrule
\end{tabular}
\end{sc} }
\end{center}
\end{table}

\begin{table}[htb] 
\caption{Description of Large Real Datasets.}
\label{dataset}
\centering\small
\begin{sc}
\resizebox{\columnwidth}{!}{
\begin{tabular}{lccccr}
\toprule
Dataset & \#Train& \#Test & \#Dimensions & Field\\
\midrule
  cadata  & 18,432 & 2,208 &8 &Housing \\
Proteins&40,730  &5,000& 9 &Biomedicine \\
APS Failure &60,000  & 16,000 &171& Vehicle \\
 MSD   &463,715 &51,630& 90&Music \\
  Gas Sensor   &4,095,000 &900,900& 20&Chemistry \\
\bottomrule
\end{tabular}}
\end{sc}  
\end{table}

\paragraph{Platforms} We develop the library in C++ and our experiments are performed on a high-performance cluster. Each machine of the cluster is composed of two Intel Xeon Platinum 9242 processors with 2.30 GHz clock speed (turbo boost frequency of up to 3.80 GHz), which owns
96 physical cores organized into two sockets. The processor contains a 71.5 MB smart cache. AVX512 instruction set extension is supported and it’s able to conduct operations for 8 double-precision floating-point data in a SIMD manner.

\paragraph{Baselines} The experiments are conducted in two parts. First, the proposed Best Friend Clustering is evaluated with three classic 2D datasets. Results of K\-Means~\cite{you2018accurate}, AHC~\cite{5372757} and MSTC~\cite{li2019scaled} methods are also presented as different baselines for comparison. 
Then we turn to the evaluation on convergence, accuracy, and scalability for our regression library with five real large-scale datasets. Since DCKRR~\cite{zhang2015divide} and BKRR2~\cite{you2020fast} are two closely related papers, they are employed as two baselines in this paper. Moreover, we alternate the clustering methods in BKRR2 with AHC~\cite{5372757} and MSTC~\cite{li2019scaled} as AHCKRR and MSTKRR respectively. Thus the experimental configurations for parallel regression cover the whole spectrum of representative clustering methods: BFCKRR (Best Friend Clustering), BKRR2~(K-Means), AHCKRR~(Agglomerative Hierarchical Clustering), and MSTKRR~(Minimum Spanning Tree Clustering).

\paragraph{Datasets} Three classic shape datasets are employed to demonstrate the quality of clustering in Table~\ref{shdataset}, which can be obtained in the Clustering Basic Benchmark (CBB)~\cite{franti2018k}. They represent well-understood clustering problems and are widely-used benchmarks for checking the applicability of clustering algorithms~\cite{zahn1971graph,gionis2007clustering,veenman2002maximum,franti2018k}. Then Datasets Million Song Data~(MSD) and Cadata are used as two of our evaluated datasets for regression since they are both used in the paper of DCKRR and BKRR2~\cite{you2020fast}. 
To further justify the scaling efficiency of our approach, we use another three real datasets, which contain the data on higher dimensions and interdisciplinary research. The details of these five datasets are sorted by \#Train and summarized in Table~\ref{dataset}. All these datasets are available in the UCI Machine Learning Repository~\cite{Dua:2019}. The distance measure is also adopted fairly by using Euclidean distance.
%

\subsection{Visualization}
\begin{figure*}
\begin{center}
\centerline{\includegraphics[width=17cm]{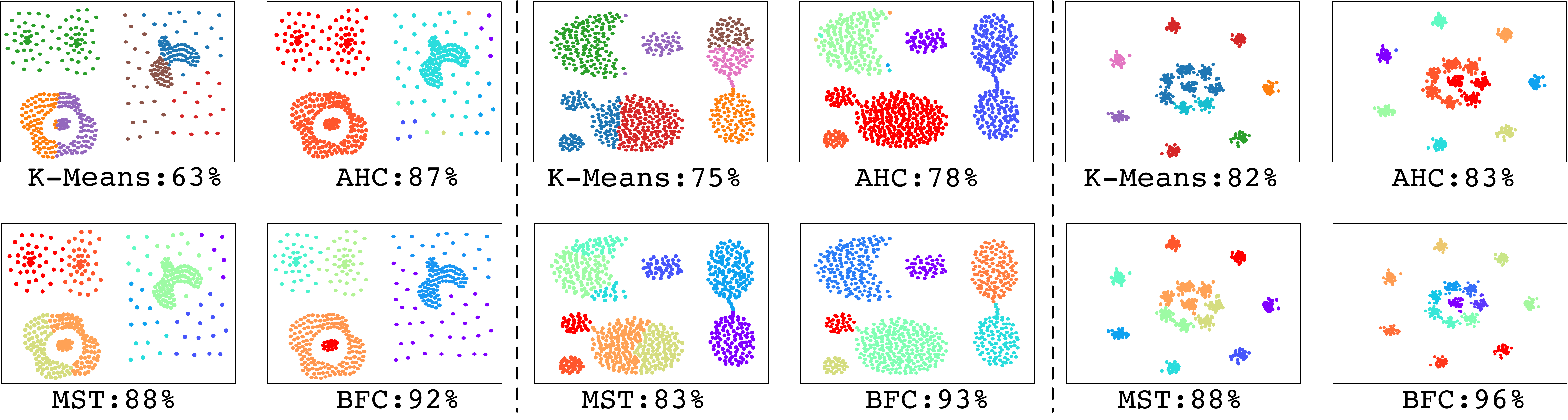}}
\caption{Visualization for four considered baselines on Zahn's Compound~\cite{zahn1971graph}~(left), Aggregation~\cite{gionis2007clustering}~(Center), and R15~\cite{veenman2002maximum}~(right) with AMI score annotated.}
\label{visu}
\end{center}
\end{figure*}

For better clarity, the quality of Best friend Clustering is visualized intuitively in Figure~\ref{visu} by utilizing the classic shape datasets in Table~\ref{shdataset}. To evaluate the classification performance quantitatively, the adjusted mutual information (AMI)~\cite{vinh2009information} is used in Figure~\ref{visu} to measure the similarity between  partitions on ground truth data and the cluster assignment obtained by the method. Due to that the performance of K-means is greatly influenced by the $K$ value, we set $K$ with real \#labels straightway. Nevertheless, K\-means yields poor results especially on Zahn's Compound~\cite{zahn1971graph} and it cannot separate the arbitrary shapes smoothly. AHC and MST clustering outperform K-means in most cases. Compared with three methods mentioned above, we can observe that BFC maintains the merges quite well and clusters these data better than the considered baselines. 

\subsection{Convergence}

\begin{table}[!t]
  \centering\small
  \resizebox{8cm}{!}{
  \setlength{\tabcolsep}{2.5mm}{
  \begin{sc}
  \begin{threeparttable}
  \setlength{\abovecaptionskip}{0.2cm} 
  \caption{\label{tab:hci}Hierarchical HCIs in Best Friend Clustering}
  \begin{tabular}{lccccccccc}
  \toprule  
  Steps &1&2&3&4&5&6&7\\\midrule
 Cadata& 0.31&\cellcolor{lightgray} 0.51 & 0.26 & 0.45&0.24&0&-\\
 Proteins&\cellcolor{lightgray}0.65& 0.44 & 0.28& 0.14&0.13&0.11&0 \\
 APS Failure& 0.30& 0.56& 0.27& 0.35 &\cellcolor{lightgray}0.58&0.50&0\\
 MSD& 0.09 &\cellcolor{lightgray}0.25 & 0.19 & 0.02&0&-&- \\ 
 Gas Sensor& 0.22 &0.38 &\cellcolor{lightgray}0.62 & 0.24&0.17&0.13&0 \\ 
  \bottomrule \end{tabular} \small
  \end{threeparttable}
  \end{sc}
  }}
\end{table}

\begin{table}[!t]
  \centering\small
  \resizebox{\columnwidth}{!}{
  \begin{sc}
  \begin{threeparttable}
  \setlength{\abovecaptionskip}{0.2cm} 
  \caption{\label{tab:clusters}\#clusters for each hierarchy in Best Friend Clustering}
  \begin{tabular}{lccccccccc}
  \toprule  
  Steps &0&1&2&3&4&5&6&7\\\midrule
 Cadata& 18,432& 4,737 & \cellcolor{lightgray}326 & 64&8&2&1&-\\
 Proteins& 40,730 & \cellcolor{lightgray}11,273 & 724& 191&32&6&2&1 \\
 APS Failure& 60,000& 15,230& 712& 124 &11&\cellcolor{lightgray}4&2&1\\
 MSD& 463,715 &54,307 & \cellcolor{lightgray}1,023 & 60&7&1&-&- \\ 
Gas Sensor&4,095,000 &231,541 &13,115 & \cellcolor{lightgray}1,721&202&10&2&1 \\
  \bottomrule \end{tabular} \small
  \end{threeparttable} 
  \end{sc}}
  \vskip -0.12in
\end{table}

Since the connected samples generated by our clustering method are grouped together, BFC could reach a fast convergence in O(log$N$) rounds. The details for HCIs and \#clusters are given in Table~\ref{tab:hci} and Table~\ref{tab:clusters} respectively, where the best convergent hierarchies are highlighted. The average sizes for each cluster are 56, 4, 15,000, 453, and 2,379 respectively on five datasets. 
Moreover, the number of iterations still falls in low single digits on large-scale datasets like Gas Sensor, which illustrates that the clustering is less sensitive to the increasing training size and appropriate for scaling cases.

\subsection{Accuracy}
\begin{figure*}
\begin{center}
\centerline{\includegraphics[width=18cm]{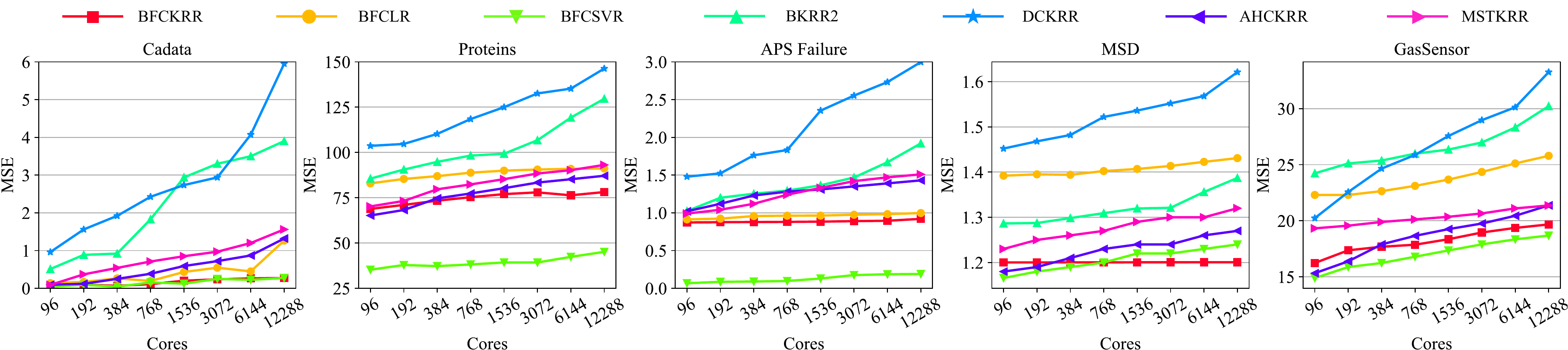}}
\caption{MSE comparison for different methods on real datasets. Our parallel library includes KRR, LR, and SVR techniques. They are applied with BFC method and abbreviated to BFCKRR, BFCLR, and BFCSVR respectively. Baselines for parallel regression applied with representative clustering methods are DCKRR~(Divide-and-Conquer without clustering)~\cite{zhang2015divide}, BKRR2~(K-Means)~\cite{you2020fast}, AHCKRR~(Agglomerative Hierarchical Clustering)~\cite{5372757} and MSTKRR(Minimum Spanning Tree Clustering)~\cite{li2019scaled}. }
\label{accuracy}
\end{center}
\end{figure*}

The proposed balanced partition strategy has little impact on the
accuracy of the model since it reserves the samples with most similarities in same cluster. Nonetheless, to assure the
correctness of our implementation, we perform the configurations scaling from 96 to 12,288 cores and compare with the reported accuracy (measured by MSE).
To give a fair comparison, best parameters were finely tuned from the same parameter set to achieve the lowest MSE in different methods. As shown in Figure~\ref{accuracy}, BFCKRR achieves the lowest MSEs when KRR techniques are employed. Moreover, BFCLR and BFCSVR could also make a superior prediction in most cases. Among all considered baselines, DCKRR and BKRR2 produce a poor quality, which adopts no samples clustering and K-means clustering respectively. As the core increases, high accuracy is obtained steadily by our library while the MSEs of other baselines ripples drastically. This illustrates that our balanced partition algorithm adds great support to accuracy in scaling cases.

\subsection{Scalability}

\begin{figure*}[htb]
\begin{center}
\centerline{\includegraphics[width=18cm]{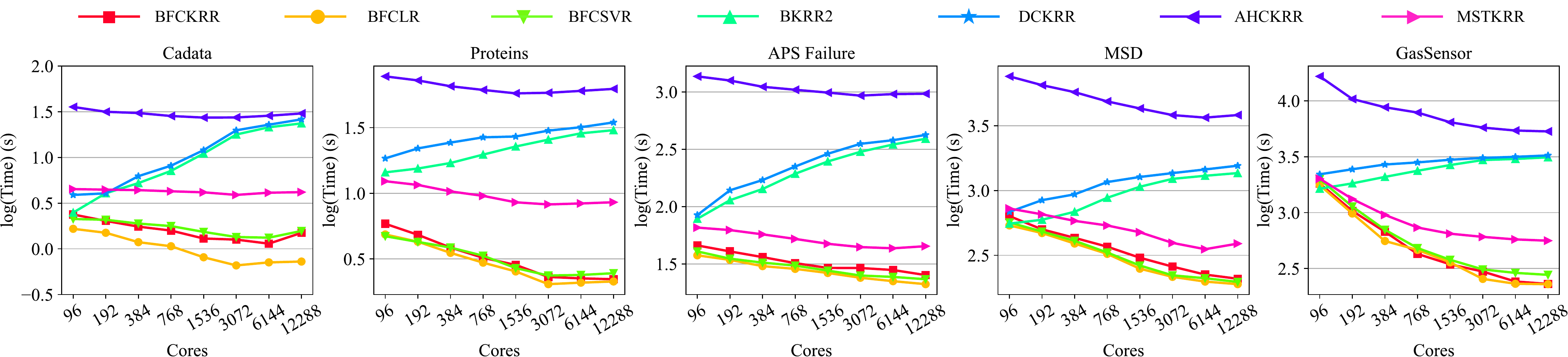}}
\caption{Scalability for different methods on five datasets. Our parallel library includes KRR, LR, and SVR techniques. They are applied with BFC method and abbreviated to BFCKRR, BFCLR, and BFCSVR respectively. Baselines for parallel regression applied with representative clustering methods are DCKRR~(Divide-and-Conquer without clustering)~\cite{zhang2015divide}, BKRR2~(K-Means)~\cite{you2020fast}, AHCKRR~(Agglomerative Hierarchical Clustering)~\cite{5372757} and MSTKRR(Minimum Spanning Tree Clustering)~\cite{li2019scaled}. }
\label{scalings}
\end{center}
\vskip -0.1in
\end{figure*}

Figure~\ref{scalings} illustrates the scalability for different methods on five datasets. We observe that our library consistently achieves high performance on them compared to baselines. With a larger dataset like Gas Sensor, gaps between them are further widened and it is even more than 13.6x faster than BKRR2. Moreover, the time for our library decreases regularly as we double the number of cores, while both DCKRR and BKRR2 exhibit a bad scaling performance by a growing curve. This illustrates that the K-means-based or DC-based methods are poor in large-scale regression, where performances are jeopardized cumulatively by the increasing $K$ value and expensive DC operations. With similar parallel implementation achieved, MSTKRR outperforms BKRR2 in most cases. However, AHCKRR suffers from the lowest performance as agglomerative style algorithms come with a quadratic time complexity inherently.

\begin{figure}[htb]
\begin{center}
\centerline{\includegraphics[width=\columnwidth]{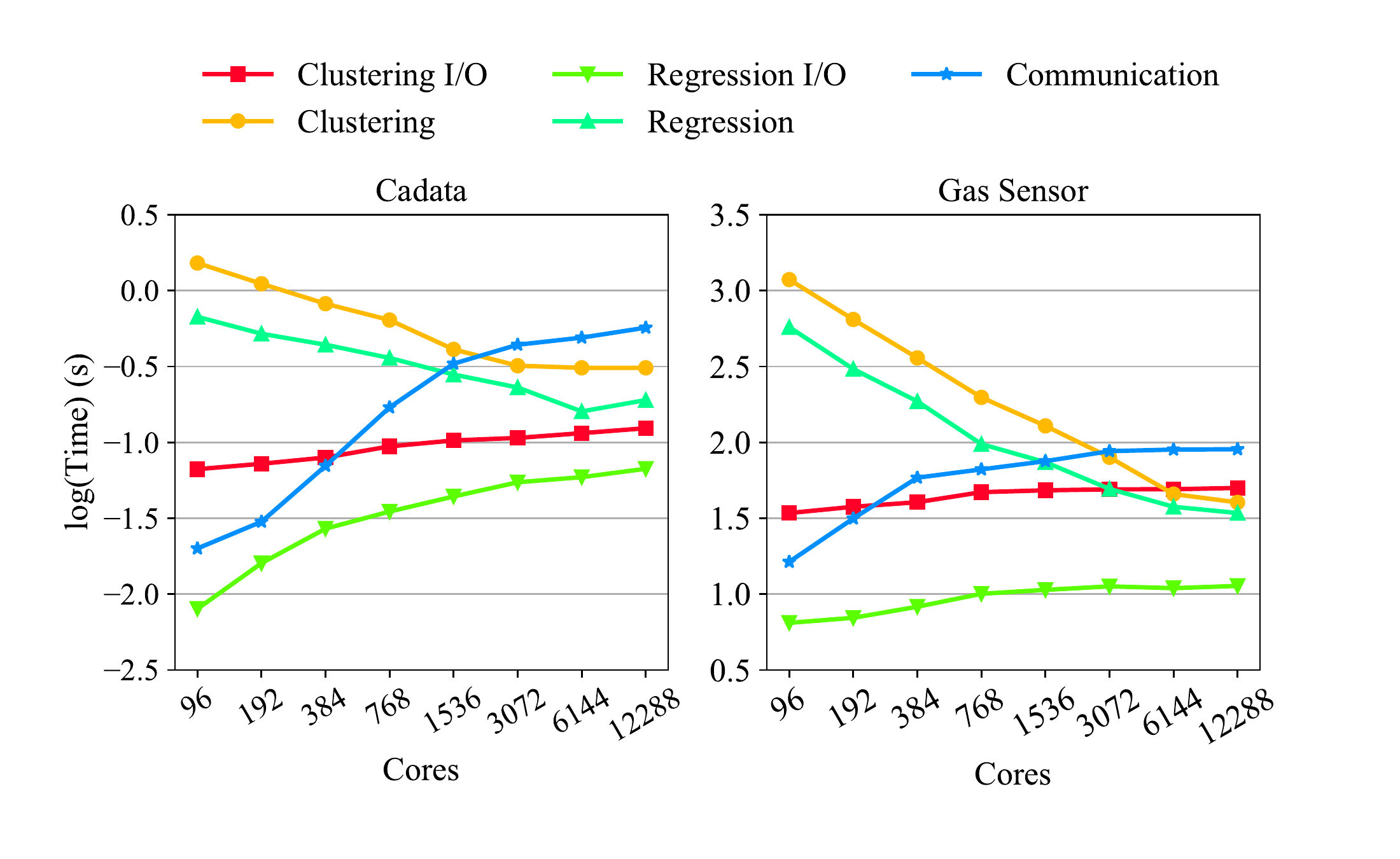}}
\vskip -0.1in
\caption{Time dissection on Cadata \& Gas Sensor datasets.}
\label{analysis}
\end{center}
\vskip -0.2in
\end{figure}

To dissect the procedures of our library and see how it varies as the size of datasets grows, we provide a quantitative look into the cases of two typical datasets, i.e., the smallest and largest ones. Figure~\ref{analysis} compares the logarithmic time for clustering I/O, clustering, regression I/O, regression, and communication in our library. Upon inspection it becomes distinct that the increasing cores also aggravate the communication and I/O cost. Despite the scaling pressure brought by communication and I/O, our library still obtains a sustained scaling performance. Table~\ref{tab:analysis} shows the analytical proportion and speedup of two datasets. Interestingly, although the average proportion on different datasets contains little difference, our library can obtain a higher speedup on Gas Sensor. Based on conjoint analyses on Figure~\ref{analysis}, we can observe that the key contribution lies in the better scaling efficiency of parallelizable procedures~(clustering and regression parts) on large-scale datasets.

\begin{table*}[] 
\centering\small
\resizebox{\textwidth}{!}{
\begin{sc}
\begin{threeparttable}
\setlength{\abovecaptionskip}{0.2cm}
\caption{\label{tab:analysis}Analytical Proportion and speedup for different procedures by our library on Cadata and Gas Sensor datasets.}
\begin{tabular}{lcccccccccccccr}
\toprule
Dataset   & \multicolumn{7}{c}{Cadata}               & \multicolumn{7}{c}{Gas Sensor}           \\
\cmidrule(r){0-0}\cmidrule(r){2-8}\cmidrule(r){9-15}
Component & C.I/O(\%)$^1$ & C.(\%) & R.I/O(\%) & R.(\%) & M.(\%) & S.(C.+R.)$^2$ & S. & C.I/O(\%) & C.(\%) & R.I/O(\%) & R.(\%) & M.(\%) & S.(C.+R.) & S. \\
\cmidrule(r){0-0}\cmidrule(r){2-2}\cmidrule(r){3-3}\cmidrule(r){4-4}\cmidrule(r){5-5}\cmidrule(r){6-6}\cmidrule(r){7-7}\cmidrule(r){8-8}\cmidrule(r){9-9}\cmidrule(r){10-10}\cmidrule(r){11-11}\cmidrule(r){12-12}\cmidrule(r){13-13}\cmidrule(r){14-14}\cmidrule(r){15-15}\\
96        & 2.80 & 63.98 & 0.34 & 28.28 & 0.84  & 1.00 & 1.00 & 1.88 & 64.84 & 0.35 & 31.51 & 0.89  & 1.00 & 1.00 \\
192       & 3.58 & 54.96 & 0.54 & 25.75 & 1.49  & 1.34 & 1.18 & 3.64 & 62.33 & 0.67 & 29.38 & 3.04  & 1.85 & 1.76 \\
384       & 4.54 & 46.88 & 1.54 & 25.15 & 4.00  & 1.74 & 1.36 & 5.97 & 53.34 & 1.22 & 27.59 & 8.67  & 3.21 & 2.70 \\
768       & 5.94 & 40.41 & 2.21 & 22.73 & 10.73 & 2.19 & 1.50 & 11.00    & 46.31 & 2.35 & 22.80 & 15.53 & 5.95 & 4.27 \\
1536      & 7.94 & 31.62 & 3.39 & 21.60 & 25.45 & 3.18 & 1.83 & 14.00    & 37.21 & 3.09 & 21.45 & 21.79 & 8.68 & 5.28 \\
3072      & 8.48 & 25.35 & 4.33 & 18.22 & 34.86 & 3.99 & 1.88 & 16.50    & 26.83 & 3.78 & 16.60 & 29.45 & 13.61    & 6.14 \\
6144      & 10.09    & 27.21 & 5.18 & 14.04 & 43.01 & 4.66 & 2.09 & 20.27    & 18.84 & 4.52 & 15.54 & 37.01 & 21.09    & 7.53 \\
12288     & 8.29 & 20.73 & 4.48 & 12.70 & 39.45 & 4.38 & 1.59 & 21.75    & 17.46 & 4.92 & 14.86 & 39.12 & 23.60    & 7.92 \\

\cmidrule(r){0-0}\cmidrule(r){2-2}\cmidrule(r){3-3}\cmidrule(r){4-4}\cmidrule(r){5-5}\cmidrule(r){6-6}\cmidrule(r){7-7}\cmidrule(r){8-8}\cmidrule(r){9-9}\cmidrule(r){10-10}\cmidrule(r){11-11}\cmidrule(r){12-12}\cmidrule(r){13-13}\cmidrule(r){14-14}\cmidrule(r){15-15}
Mean    & 6.46    & 38.89 & 2.75    & 21.06 & 19.98 & 2.81    & 1.55        & 11.88   & 40.90 & 2.61    & 22.47 & 19.44 & 9.87    & 4.57 \\\bottomrule
\end{tabular}
\begin{tablenotes}
    \item[1] For better clarity, procedures for clustering, regression, and communication are abbreviated with C., R., and M. respectively. 
    \item[2] The speedup is also abbreviated to S..
\end{tablenotes}
\end{threeparttable}
\end{sc} 
}
\end{table*}

\section{\label{relatedwork}Related Work}
 
Our paper shows closely concerns for two main lines of research. 
The first thread is to study the efficient clustering methods with better-quality on distributed systems. Basically, AHC and K-Means are two extensively used  methods~\cite{tan2016introduction}. 
Bahmani~\textit{et al.}  extended the k-means problems by a MapReduce algorithm on distributed nodes~\cite{ScalableKMeans}. Ene~\textit{et al.} optimized the greedy algorithm on MapReduce to solve the k-center problems and the
local search strategy to address the k-median problems ~\cite{ene2011fast}. 
Subsequently, several papers studied similar problems with the MapReduce model~\cite{balcan2013distributed,bateni2014distributed,anchalia2013mapreduce}. Then Bouguettaya~\textit{et al.} built a hierarchy based on a group of centroids generated by K-Means to improve the efficiency of AHC~\cite{bouguettaya2015efficient}, while it only was implemented for a single node.
As for Graph Clustering, massive efforts were also put into studying efficient algorithms~\cite{ahn2012analyzing,andoni2014parallel,chitnis2016kernelization,esfandiari2018streaming}. Nevertheless the hierarchical characters attracted little attention in these work.
Grygorash~\textit{et al.} presented the Hierarchical euclidean-distance-based MST clustering algorithm
(HEMST). Given the number of clusters as an input, HEMST accelerate the convergence progress by merging multiple edges, while the edges larger than the threshold are required to be sorted and removed in order~\cite{grygorash2006minimum}. Jin~\textit{et al.} split the clustering problem into various overlapped subproblems by a Prim algorithm, solved each subproblem, and then merged them into an overall solution~\cite{jin2013disc}. However, the MSTs are needed to store at the Map side and then shuffle to the Reducers. 
Bateni~\textit{et al.}~\cite{bateni2017affinity} extended MST-based method called affinity clustering with two classic MST algorithms.
It still relied heavily on MapReduce and requires moving all the edges to one machine serially for MSTs after the edges deletion. Moreover, affinity clustering did not reveal the exact running times and number of machines used in their experiments~\cite{bateni2017affinity}. Wang~\textit{et al.} utilized a divide-and-conquer scheme to construct approximate MSTs, while the process to detect the long edges of the MST is also highly sequential at an early stage for clustering~\cite{wang2009divide,zhong2015fast}.
As a result, an efficient clustering algorithm competent for parallel computing on large-scale data is in need crucially to improve the accuracy of K-means, efficiency of AHC, and scalability of GC methods.

The second focus of this paper is closely related to distributed regression, which suffers from serious scalability problems in both computation time and memory usage~\cite{chen2014data}.
Mini-batch Gradient Descent (MBGD) was used for training on  batched data~\cite{hinton2012neural}, while it was proposed for serial implementation. MapReduce-based methods~\cite{chu2007map,teo2010bundle,he2013parallel,yang2012mapreduce} made computation distributed locally which holds parts of the data. However, the overall cost in terms of computation and network is high because of the busy synchronization sweeps.
The Divide-and-Conquer algorithm was then adopted on distributed systems for SVM and KRR~\cite{hsieh2014divide,zhang2013divide,zhang2015divide}.
Parallel SVM (PSVM) is presented recently to decrease memory and time consumption~\cite{dong2005fast,mitra2004probabilistic,bekkerman2011scaling}.
Zhang~\textit{et al.}~proved that regression with kernel method is more accurate than non-kernel methods, and DCKRR designed by them can outperform all the previous approximate methods~\cite{zhang2013divide}. 
You~\textit{et al.}~\cite{you2018accurate} presented K-Means kernel Ridge Regression (KKRR) for efficient regression on clustered data.
Recent work BKRR2~\cite{you2020fast} was optimized on KKRR by averaging the loads on each node and had better accuracy than DCKRR and KKRR, which was considered as a state-of-the-art approach for parallel kernel regression. Thus, in this paper, the focus is paid to the comparison with two closely related papers DCKRR and BKRR2, which are also elaborated in the Introduction section.

\section{Conclusion}
In this paper, we first propose a Best Friend Clustering method, which is more accurate, fast, and meanwhile parameter-free. Then we devise a strategy to determine the optimal aggregation through a well-defined metric and achieve it efficiently on distributed systems. 
Moreover, a balanced partition algorithm inspired by backtracking is devised for load-balance in parallel implementation. At last, we integrate the proposed methods into multiple regression techniques for a parallel library, and it shows  superior performance on convergence, accuracy, and scalability.


\bibliographystyle{ACM-Reference-Format} 
\bibliography{example_paper} 


\begin{thebibliography}{54}


\ifx \showCODEN    \undefined \def \showCODEN     #1{\unskip}     \fi
\ifx \showDOI      \undefined \def \showDOI       #1{#1}\fi
\ifx \showISBNx    \undefined \def \showISBNx     #1{\unskip}     \fi
\ifx \showISBNxiii \undefined \def \showISBNxiii  #1{\unskip}     \fi
\ifx \showISSN     \undefined \def \showISSN      #1{\unskip}     \fi
\ifx \showLCCN     \undefined \def \showLCCN      #1{\unskip}     \fi
\ifx \shownote     \undefined \def \shownote      #1{#1}          \fi
\ifx \showarticletitle \undefined \def \showarticletitle #1{#1}   \fi
\ifx \showURL      \undefined \def \showURL       {\relax}        \fi
\providecommand\bibfield[2]{#2}
\providecommand\bibinfo[2]{#2}
\providecommand\natexlab[1]{#1}
\providecommand\showeprint[2][]{arXiv:#2}

\bibitem[\protect\citeauthoryear{Ahn, Guha, and McGregor}{Ahn
  et~al\mbox{.}}{2012}]%
        {ahn2012analyzing}
\bibfield{author}{\bibinfo{person}{Kook~Jin Ahn}, \bibinfo{person}{Sudipto
  Guha}, {and} \bibinfo{person}{Andrew McGregor}.}
  \bibinfo{year}{2012}\natexlab{}.
\newblock \showarticletitle{Analyzing graph structure via linear measurements}.
  In \bibinfo{booktitle}{\emph{Proceedings of the twenty-third annual ACM-SIAM
  symposium on Discrete Algorithms}}. SIAM, \bibinfo{pages}{459--467}.
\newblock


\bibitem[\protect\citeauthoryear{Anchalia, Koundinya, and Srinath}{Anchalia
  et~al\mbox{.}}{2013}]%
        {anchalia2013mapreduce}
\bibfield{author}{\bibinfo{person}{Prajesh~P Anchalia},
  \bibinfo{person}{Anjan~K Koundinya}, {and} \bibinfo{person}{NK Srinath}.}
  \bibinfo{year}{2013}\natexlab{}.
\newblock \showarticletitle{MapReduce design of K-means clustering algorithm}.
  In \bibinfo{booktitle}{\emph{2013 International Conference on Information
  Science and Applications (ICISA)}}. IEEE, \bibinfo{pages}{1--5}.
\newblock


\bibitem[\protect\citeauthoryear{Andoni, Nikolov, Onak, and
  Yaroslavtsev}{Andoni et~al\mbox{.}}{2014}]%
        {andoni2014parallel}
\bibfield{author}{\bibinfo{person}{Alexandr Andoni},
  \bibinfo{person}{Aleksandar Nikolov}, \bibinfo{person}{Krzysztof Onak}, {and}
  \bibinfo{person}{Grigory Yaroslavtsev}.} \bibinfo{year}{2014}\natexlab{}.
\newblock \showarticletitle{Parallel algorithms for geometric graph problems}.
  In \bibinfo{booktitle}{\emph{Proceedings of the forty-sixth annual ACM
  symposium on Theory of computing}}. \bibinfo{pages}{574--583}.
\newblock


\bibitem[\protect\citeauthoryear{Asano, Bhattacharya, Keil, and Yao}{Asano
  et~al\mbox{.}}{1988}]%
        {asano1988clustering}
\bibfield{author}{\bibinfo{person}{Tetsuo Asano}, \bibinfo{person}{Binay
  Bhattacharya}, \bibinfo{person}{Mark Keil}, {and} \bibinfo{person}{Frances
  Yao}.} \bibinfo{year}{1988}\natexlab{}.
\newblock \showarticletitle{Clustering algorithms based on minimum and maximum
  spanning trees}. In \bibinfo{booktitle}{\emph{Proceedings of the fourth
  annual symposium on Computational geometry}}. \bibinfo{pages}{252--257}.
\newblock


\bibitem[\protect\citeauthoryear{Bahmani, Moseley, Vattani, Kumar, and
  Vassilvitskii}{Bahmani et~al\mbox{.}}{2012}]%
        {ScalableKMeans}
\bibfield{author}{\bibinfo{person}{Bahman Bahmani}, \bibinfo{person}{Benjamin
  Moseley}, \bibinfo{person}{Andrea Vattani}, \bibinfo{person}{Ravi Kumar},
  {and} \bibinfo{person}{Sergei Vassilvitskii}.}
  \bibinfo{year}{2012}\natexlab{}.
\newblock \showarticletitle{Scalable K-Means++}.
\newblock \bibinfo{journal}{\emph{Proc. VLDB Endow.}} \bibinfo{volume}{5},
  \bibinfo{number}{7} (\bibinfo{date}{March} \bibinfo{year}{2012}),
  \bibinfo{pages}{622–633}.
\newblock
\showISSN{2150-8097}
\urldef\tempurl%
\url{https://doi.org/10.14778/2180912.2180915}
\showDOI{\tempurl}


\bibitem[\protect\citeauthoryear{Balcan, Ehrlich, and Liang}{Balcan
  et~al\mbox{.}}{2013}]%
        {balcan2013distributed}
\bibfield{author}{\bibinfo{person}{Maria-Florina~F Balcan},
  \bibinfo{person}{Steven Ehrlich}, {and} \bibinfo{person}{Yingyu Liang}.}
  \bibinfo{year}{2013}\natexlab{}.
\newblock \showarticletitle{Distributed $ k $-means and $ k $-median Clustering
  on General Topologies}.
\newblock \bibinfo{journal}{\emph{Advances in Neural Information Processing
  Systems}}  \bibinfo{volume}{26} (\bibinfo{year}{2013}),
  \bibinfo{pages}{1995--2003}.
\newblock


\bibitem[\protect\citeauthoryear{Bateni, Bhaskara, Lattanzi, and
  Mirrokni}{Bateni et~al\mbox{.}}{2014}]%
        {bateni2014distributed}
\bibfield{author}{\bibinfo{person}{MohammadHossein Bateni},
  \bibinfo{person}{Aditya Bhaskara}, \bibinfo{person}{Silvio Lattanzi}, {and}
  \bibinfo{person}{Vahab~S Mirrokni}.} \bibinfo{year}{2014}\natexlab{}.
\newblock \showarticletitle{Distributed Balanced Clustering via Mapping
  Coresets.}. In \bibinfo{booktitle}{\emph{NIPS}}. \bibinfo{pages}{2591--2599}.
\newblock


\bibitem[\protect\citeauthoryear{Bateni, Behnezhad, Derakhshan, Hajiaghayi,
  Kiveris, Lattanzi, and Mirrokni}{Bateni et~al\mbox{.}}{2017}]%
        {bateni2017affinity}
\bibfield{author}{\bibinfo{person}{Mohammad~Hossein Bateni},
  \bibinfo{person}{Soheil Behnezhad}, \bibinfo{person}{Mahsa Derakhshan},
  \bibinfo{person}{Mohammad~Taghi Hajiaghayi}, \bibinfo{person}{Raimondas
  Kiveris}, \bibinfo{person}{Silvio Lattanzi}, {and} \bibinfo{person}{Vahab
  Mirrokni}.} \bibinfo{year}{2017}\natexlab{}.
\newblock \showarticletitle{Affinity clustering: Hierarchical clustering at
  scale}. In \bibinfo{booktitle}{\emph{Proceedings of the 31st International
  Conference on Neural Information Processing Systems}}.
  \bibinfo{pages}{6867--6877}.
\newblock


\bibitem[\protect\citeauthoryear{Bekkerman, Bilenko, and Langford}{Bekkerman
  et~al\mbox{.}}{2011}]%
        {bekkerman2011scaling}
\bibfield{author}{\bibinfo{person}{Ron Bekkerman}, \bibinfo{person}{Mikhail
  Bilenko}, {and} \bibinfo{person}{John Langford}.}
  \bibinfo{year}{2011}\natexlab{}.
\newblock \bibinfo{booktitle}{\emph{Scaling up machine learning: Parallel and
  distributed approaches}}.
\newblock \bibinfo{publisher}{Cambridge University Press}.
\newblock


\bibitem[\protect\citeauthoryear{Bouguettaya, Yu, Liu, Zhou, and
  Song}{Bouguettaya et~al\mbox{.}}{2015}]%
        {bouguettaya2015efficient}
\bibfield{author}{\bibinfo{person}{Athman Bouguettaya}, \bibinfo{person}{Qi
  Yu}, \bibinfo{person}{Xumin Liu}, \bibinfo{person}{Xiangmin Zhou}, {and}
  \bibinfo{person}{Andy Song}.} \bibinfo{year}{2015}\natexlab{}.
\newblock \showarticletitle{Efficient agglomerative hierarchical clustering}.
\newblock \bibinfo{journal}{\emph{Expert Systems with Applications}}
  \bibinfo{volume}{42}, \bibinfo{number}{5} (\bibinfo{year}{2015}),
  \bibinfo{pages}{2785--2797}.
\newblock


\bibitem[\protect\citeauthoryear{Chen and Zhang}{Chen and Zhang}{2014}]%
        {chen2014data}
\bibfield{author}{\bibinfo{person}{CL~Philip Chen} {and}
  \bibinfo{person}{Chun-Yang Zhang}.} \bibinfo{year}{2014}\natexlab{}.
\newblock \showarticletitle{Data-intensive applications, challenges, techniques
  and technologies: A survey on Big Data}.
\newblock \bibinfo{journal}{\emph{Information sciences}}  \bibinfo{volume}{275}
  (\bibinfo{year}{2014}), \bibinfo{pages}{314--347}.
\newblock


\bibitem[\protect\citeauthoryear{Chen, Song, Zeng, and Jiang}{Chen
  et~al\mbox{.}}{2020}]%
        {chen2020scene}
\bibfield{author}{\bibinfo{person}{Gongwei Chen}, \bibinfo{person}{Xinhang
  Song}, \bibinfo{person}{Haitao Zeng}, {and} \bibinfo{person}{Shuqiang
  Jiang}.} \bibinfo{year}{2020}\natexlab{}.
\newblock \showarticletitle{Scene recognition with prototype-agnostic scene
  layout}.
\newblock \bibinfo{journal}{\emph{IEEE Transactions on Image Processing}}
  \bibinfo{volume}{29} (\bibinfo{year}{2020}), \bibinfo{pages}{5877--5888}.
\newblock


\bibitem[\protect\citeauthoryear{Chitnis, Cormode, Esfandiari, Hajiaghayi,
  McGregor, Monemizadeh, and Vorotnikova}{Chitnis et~al\mbox{.}}{2016}]%
        {chitnis2016kernelization}
\bibfield{author}{\bibinfo{person}{Rajesh Chitnis}, \bibinfo{person}{Graham
  Cormode}, \bibinfo{person}{Hossein Esfandiari},
  \bibinfo{person}{MohammadTaghi Hajiaghayi}, \bibinfo{person}{Andrew
  McGregor}, \bibinfo{person}{Morteza Monemizadeh}, {and}
  \bibinfo{person}{Sofya Vorotnikova}.} \bibinfo{year}{2016}\natexlab{}.
\newblock \showarticletitle{Kernelization via sampling with applications to
  finding matchings and related problems in dynamic graph streams}. In
  \bibinfo{booktitle}{\emph{Proceedings of the twenty-seventh annual ACM-SIAM
  symposium on Discrete algorithms}}. SIAM, \bibinfo{pages}{1326--1344}.
\newblock


\bibitem[\protect\citeauthoryear{Chu, Kim, Lin, Yu, Bradski, Olukotun, and
  Ng}{Chu et~al\mbox{.}}{2007}]%
        {chu2007map}
\bibfield{author}{\bibinfo{person}{Cheng-Tao Chu}, \bibinfo{person}{Sang~K
  Kim}, \bibinfo{person}{Yi-An Lin}, \bibinfo{person}{YuanYuan Yu},
  \bibinfo{person}{Gary Bradski}, \bibinfo{person}{Kunle Olukotun}, {and}
  \bibinfo{person}{Andrew~Y Ng}.} \bibinfo{year}{2007}\natexlab{}.
\newblock \showarticletitle{Map-reduce for machine learning on multicore}. In
  \bibinfo{booktitle}{\emph{Advances in neural information processing
  systems}}. \bibinfo{pages}{281--288}.
\newblock


\bibitem[\protect\citeauthoryear{Day and Edelsbrunner}{Day and
  Edelsbrunner}{1984}]%
        {day1984efficient}
\bibfield{author}{\bibinfo{person}{William~HE Day} {and}
  \bibinfo{person}{Herbert Edelsbrunner}.} \bibinfo{year}{1984}\natexlab{}.
\newblock \showarticletitle{Efficient algorithms for agglomerative hierarchical
  clustering methods}.
\newblock \bibinfo{journal}{\emph{Journal of classification}}
  \bibinfo{volume}{1}, \bibinfo{number}{1} (\bibinfo{year}{1984}),
  \bibinfo{pages}{7--24}.
\newblock


\bibitem[\protect\citeauthoryear{Dhillon, Guan, and Kulis}{Dhillon
  et~al\mbox{.}}{2004}]%
        {dhillon2004kernel}
\bibfield{author}{\bibinfo{person}{Inderjit~S Dhillon},
  \bibinfo{person}{Yuqiang Guan}, {and} \bibinfo{person}{Brian Kulis}.}
  \bibinfo{year}{2004}\natexlab{}.
\newblock \showarticletitle{Kernel k-means: spectral clustering and normalized
  cuts}. In \bibinfo{booktitle}{\emph{Proceedings of the tenth ACM SIGKDD
  international conference on Knowledge discovery and data mining}}.
  \bibinfo{pages}{551--556}.
\newblock


\bibitem[\protect\citeauthoryear{Ding and He}{Ding and He}{2004}]%
        {ding2004k}
\bibfield{author}{\bibinfo{person}{Chris Ding} {and} \bibinfo{person}{Xiaofeng
  He}.} \bibinfo{year}{2004}\natexlab{}.
\newblock \showarticletitle{K-means clustering via principal component
  analysis}. In \bibinfo{booktitle}{\emph{Proceedings of the twenty-first
  international conference on Machine learning}}. \bibinfo{pages}{29}.
\newblock


\bibitem[\protect\citeauthoryear{Dong, Krzyzak, and Suen}{Dong
  et~al\mbox{.}}{2005}]%
        {dong2005fast}
\bibfield{author}{\bibinfo{person}{Jian-xiong Dong}, \bibinfo{person}{Adam
  Krzyzak}, {and} \bibinfo{person}{Ching~Y Suen}.}
  \bibinfo{year}{2005}\natexlab{}.
\newblock \showarticletitle{Fast SVM training algorithm with decomposition on
  very large data sets}.
\newblock \bibinfo{journal}{\emph{IEEE transactions on pattern analysis and
  machine intelligence}} \bibinfo{volume}{27}, \bibinfo{number}{4}
  (\bibinfo{year}{2005}), \bibinfo{pages}{603--618}.
\newblock


\bibitem[\protect\citeauthoryear{Draper and Smith}{Draper and Smith}{1998}]%
        {draper1998applied}
\bibfield{author}{\bibinfo{person}{Norman~R Draper} {and}
  \bibinfo{person}{Harry Smith}.} \bibinfo{year}{1998}\natexlab{}.
\newblock \bibinfo{booktitle}{\emph{Applied regression analysis}}.
  Vol.~\bibinfo{volume}{326}.
\newblock \bibinfo{publisher}{John Wiley \& Sons}.
\newblock


\bibitem[\protect\citeauthoryear{Dua and Graff}{Dua and Graff}{2017}]%
        {Dua:2019}
\bibfield{author}{\bibinfo{person}{Dheeru Dua} {and} \bibinfo{person}{Casey
  Graff}.} \bibinfo{year}{2017}\natexlab{}.
\newblock \bibinfo{title}{UCI Machine Learning Repository}.
\newblock
\newblock
\urldef\tempurl%
\url{http://archive.ics.uci.edu/ml}
\showURL{%
\tempurl}


\bibitem[\protect\citeauthoryear{Ene, Im, and Moseley}{Ene
  et~al\mbox{.}}{2011}]%
        {ene2011fast}
\bibfield{author}{\bibinfo{person}{Alina Ene}, \bibinfo{person}{Sungjin Im},
  {and} \bibinfo{person}{Benjamin Moseley}.} \bibinfo{year}{2011}\natexlab{}.
\newblock \showarticletitle{Fast clustering using MapReduce}. In
  \bibinfo{booktitle}{\emph{Proceedings of the 17th ACM SIGKDD international
  conference on Knowledge discovery and data mining}}.
  \bibinfo{pages}{681--689}.
\newblock


\bibitem[\protect\citeauthoryear{Esfandiari, Hajiaghayi, Liaghat, Monemizadeh,
  and Onak}{Esfandiari et~al\mbox{.}}{2018}]%
        {esfandiari2018streaming}
\bibfield{author}{\bibinfo{person}{Hossein Esfandiari},
  \bibinfo{person}{Mohammadtaghi Hajiaghayi}, \bibinfo{person}{Vahid Liaghat},
  \bibinfo{person}{Morteza Monemizadeh}, {and} \bibinfo{person}{Krzysztof
  Onak}.} \bibinfo{year}{2018}\natexlab{}.
\newblock \showarticletitle{Streaming algorithms for estimating the matching
  size in planar graphs and beyond}.
\newblock \bibinfo{journal}{\emph{ACM Transactions on Algorithms (TALG)}}
  \bibinfo{volume}{14}, \bibinfo{number}{4} (\bibinfo{year}{2018}),
  \bibinfo{pages}{1--23}.
\newblock


\bibitem[\protect\citeauthoryear{Fr{\"a}nti and Sieranoja}{Fr{\"a}nti and
  Sieranoja}{2018}]%
        {franti2018k}
\bibfield{author}{\bibinfo{person}{Pasi Fr{\"a}nti} {and} \bibinfo{person}{Sami
  Sieranoja}.} \bibinfo{year}{2018}\natexlab{}.
\newblock \showarticletitle{K-means properties on six clustering benchmark
  datasets}.
\newblock \bibinfo{journal}{\emph{Applied Intelligence}} \bibinfo{volume}{48},
  \bibinfo{number}{12} (\bibinfo{year}{2018}), \bibinfo{pages}{4743--4759}.
\newblock


\bibitem[\protect\citeauthoryear{Gionis, Mannila, and Tsaparas}{Gionis
  et~al\mbox{.}}{2007}]%
        {gionis2007clustering}
\bibfield{author}{\bibinfo{person}{Aristides Gionis}, \bibinfo{person}{Heikki
  Mannila}, {and} \bibinfo{person}{Panayiotis Tsaparas}.}
  \bibinfo{year}{2007}\natexlab{}.
\newblock \showarticletitle{Clustering aggregation}.
\newblock \bibinfo{journal}{\emph{Acm transactions on knowledge discovery from
  data (tkdd)}} \bibinfo{volume}{1}, \bibinfo{number}{1}
  (\bibinfo{year}{2007}), \bibinfo{pages}{4--es}.
\newblock


\bibitem[\protect\citeauthoryear{Grygorash, Zhou, and Jorgensen}{Grygorash
  et~al\mbox{.}}{2006}]%
        {grygorash2006minimum}
\bibfield{author}{\bibinfo{person}{Oleksandr Grygorash}, \bibinfo{person}{Yan
  Zhou}, {and} \bibinfo{person}{Zach Jorgensen}.}
  \bibinfo{year}{2006}\natexlab{}.
\newblock \showarticletitle{Minimum spanning tree based clustering algorithms}.
  In \bibinfo{booktitle}{\emph{2006 18th IEEE International Conference on Tools
  with Artificial Intelligence (ICTAI'06)}}. IEEE, \bibinfo{pages}{73--81}.
\newblock


\bibitem[\protect\citeauthoryear{Guha, Rastogi, and Shim}{Guha
  et~al\mbox{.}}{1998}]%
        {guha1998cure}
\bibfield{author}{\bibinfo{person}{Sudipto Guha}, \bibinfo{person}{Rajeev
  Rastogi}, {and} \bibinfo{person}{Kyuseok Shim}.}
  \bibinfo{year}{1998}\natexlab{}.
\newblock \showarticletitle{CURE: an efficient clustering algorithm for large
  databases}.
\newblock \bibinfo{journal}{\emph{ACM Sigmod record}} \bibinfo{volume}{27},
  \bibinfo{number}{2} (\bibinfo{year}{1998}), \bibinfo{pages}{73--84}.
\newblock


\bibitem[\protect\citeauthoryear{He, Shang, Zhuang, and Shi}{He
  et~al\mbox{.}}{2013}]%
        {he2013parallel}
\bibfield{author}{\bibinfo{person}{Qing He}, \bibinfo{person}{Tianfeng Shang},
  \bibinfo{person}{Fuzhen Zhuang}, {and} \bibinfo{person}{Zhongzhi Shi}.}
  \bibinfo{year}{2013}\natexlab{}.
\newblock \showarticletitle{Parallel extreme learning machine for regression
  based on MapReduce}.
\newblock \bibinfo{journal}{\emph{Neurocomputing}}  \bibinfo{volume}{102}
  (\bibinfo{year}{2013}), \bibinfo{pages}{52--58}.
\newblock


\bibitem[\protect\citeauthoryear{Hinton, Srivastava, and Swersky}{Hinton
  et~al\mbox{.}}{2012}]%
        {hinton2012neural}
\bibfield{author}{\bibinfo{person}{Geoffrey Hinton}, \bibinfo{person}{Nitish
  Srivastava}, {and} \bibinfo{person}{Kevin Swersky}.}
  \bibinfo{year}{2012}\natexlab{}.
\newblock \showarticletitle{Neural networks for machine learning lecture 6a
  overview of mini-batch gradient descent}.
\newblock \bibinfo{journal}{\emph{Cited on}} \bibinfo{volume}{14},
  \bibinfo{number}{8} (\bibinfo{year}{2012}).
\newblock


\bibitem[\protect\citeauthoryear{Hsieh, Si, and Dhillon}{Hsieh
  et~al\mbox{.}}{2014}]%
        {hsieh2014divide}
\bibfield{author}{\bibinfo{person}{Cho-Jui Hsieh}, \bibinfo{person}{Si Si},
  {and} \bibinfo{person}{Inderjit Dhillon}.} \bibinfo{year}{2014}\natexlab{}.
\newblock \showarticletitle{A divide-and-conquer solver for kernel support
  vector machines}. In \bibinfo{booktitle}{\emph{International conference on
  machine learning}}. \bibinfo{pages}{566--574}.
\newblock


\bibitem[\protect\citeauthoryear{Jin, Patwary, Agrawal, Hendrix, Liao, and
  Choudhary}{Jin et~al\mbox{.}}{[n. d.]}]%
        {jin2013disc}
\bibfield{author}{\bibinfo{person}{Chen Jin}, \bibinfo{person}{Md~Mostofa~Ali
  Patwary}, \bibinfo{person}{Ankit Agrawal}, \bibinfo{person}{William Hendrix},
  \bibinfo{person}{Wei-keng Liao}, {and} \bibinfo{person}{Alok Choudhary}.}
  \bibinfo{year}{[n. d.]}\natexlab{}.
\newblock \showarticletitle{Disc: A distributed single-linkage hierarchical
  clustering algorithm using mapreduce}.
\newblock \bibinfo{journal}{\emph{work}}  \bibinfo{volume}{23}
  (\bibinfo{year}{[n. d.]}), \bibinfo{pages}{27}.
\newblock


\bibitem[\protect\citeauthoryear{Kanungo, Mount, Netanyahu, Piatko, Silverman,
  and Wu}{Kanungo et~al\mbox{.}}{2002}]%
        {kanungo2002efficient}
\bibfield{author}{\bibinfo{person}{Tapas Kanungo}, \bibinfo{person}{David~M
  Mount}, \bibinfo{person}{Nathan~S Netanyahu}, \bibinfo{person}{Christine~D
  Piatko}, \bibinfo{person}{Ruth Silverman}, {and} \bibinfo{person}{Angela~Y
  Wu}.} \bibinfo{year}{2002}\natexlab{}.
\newblock \showarticletitle{An efficient k-means clustering algorithm: Analysis
  and implementation}.
\newblock \bibinfo{journal}{\emph{IEEE transactions on pattern analysis and
  machine intelligence}} \bibinfo{volume}{24}, \bibinfo{number}{7}
  (\bibinfo{year}{2002}), \bibinfo{pages}{881--892}.
\newblock


\bibitem[\protect\citeauthoryear{Kleinberg and Tardos}{Kleinberg and
  Tardos}{2006}]%
        {kleinberg2006algorithm}
\bibfield{author}{\bibinfo{person}{Jon Kleinberg} {and} \bibinfo{person}{Eva
  Tardos}.} \bibinfo{year}{2006}\natexlab{}.
\newblock \bibinfo{booktitle}{\emph{Algorithm design}}.
\newblock \bibinfo{publisher}{Pearson Education India}.
\newblock


\bibitem[\protect\citeauthoryear{Kuhn}{Kuhn}{2015}]%
        {kuhn2015caret}
\bibfield{author}{\bibinfo{person}{Max Kuhn}.} \bibinfo{year}{2015}\natexlab{}.
\newblock \showarticletitle{Caret: classification and regression training}.
\newblock \bibinfo{journal}{\emph{ascl}} (\bibinfo{year}{2015}),
  \bibinfo{pages}{ascl--1505}.
\newblock


\bibitem[\protect\citeauthoryear{Li, Wang, and Wang}{Li et~al\mbox{.}}{2019b}]%
        {li2019scaled}
\bibfield{author}{\bibinfo{person}{Jia Li}, \bibinfo{person}{Xiaochun Wang},
  {and} \bibinfo{person}{Xiali Wang}.} \bibinfo{year}{2019}\natexlab{b}.
\newblock \showarticletitle{A scaled-MST-based clustering algorithm and
  application on image segmentation}.
\newblock \bibinfo{journal}{\emph{Journal of Intelligent Information Systems}}
  (\bibinfo{year}{2019}), \bibinfo{pages}{1--25}.
\newblock


\bibitem[\protect\citeauthoryear{Li, Shang, Zhang, Li, Wu, Wang, Zhang, Li,
  Chen, and Wei}{Li et~al\mbox{.}}{2019a}]%
        {li2019openkmc}
\bibfield{author}{\bibinfo{person}{Kun Li}, \bibinfo{person}{Honghui Shang},
  \bibinfo{person}{Yunquan Zhang}, \bibinfo{person}{Shigang Li},
  \bibinfo{person}{Baodong Wu}, \bibinfo{person}{Dong Wang},
  \bibinfo{person}{Libo Zhang}, \bibinfo{person}{Fang Li},
  \bibinfo{person}{Dexun Chen}, {and} \bibinfo{person}{Zhiqiang Wei}.}
  \bibinfo{year}{2019}\natexlab{a}.
\newblock \showarticletitle{OpenKMC: a KMC design for hundred-billion-atom
  simulation using millions of cores on Sunway Taihulight}. In
  \bibinfo{booktitle}{\emph{Proceedings of the International Conference for
  High Performance Computing, Networking, Storage and Analysis}}.
  \bibinfo{pages}{1--16}.
\newblock


\bibitem[\protect\citeauthoryear{Likas, Vlassis, and Verbeek}{Likas
  et~al\mbox{.}}{2003}]%
        {likas2003global}
\bibfield{author}{\bibinfo{person}{Aristidis Likas}, \bibinfo{person}{Nikos
  Vlassis}, {and} \bibinfo{person}{Jakob~J Verbeek}.}
  \bibinfo{year}{2003}\natexlab{}.
\newblock \showarticletitle{The global k-means clustering algorithm}.
\newblock \bibinfo{journal}{\emph{Pattern recognition}} \bibinfo{volume}{36},
  \bibinfo{number}{2} (\bibinfo{year}{2003}), \bibinfo{pages}{451--461}.
\newblock


\bibitem[\protect\citeauthoryear{Mitra, Murthy, and Pal}{Mitra
  et~al\mbox{.}}{2004}]%
        {mitra2004probabilistic}
\bibfield{author}{\bibinfo{person}{Pabitra Mitra}, \bibinfo{person}{CA Murthy},
  {and} \bibinfo{person}{Sankar~K Pal}.} \bibinfo{year}{2004}\natexlab{}.
\newblock \showarticletitle{A probabilistic active support vector learning
  algorithm}.
\newblock \bibinfo{journal}{\emph{IEEE Transactions on Pattern Analysis and
  Machine Intelligence}} \bibinfo{volume}{26}, \bibinfo{number}{3}
  (\bibinfo{year}{2004}), \bibinfo{pages}{413--418}.
\newblock


\bibitem[\protect\citeauthoryear{Murtagh and Contreras}{Murtagh and
  Contreras}{2012}]%
        {murtagh2012algorithms}
\bibfield{author}{\bibinfo{person}{Fionn Murtagh} {and} \bibinfo{person}{Pedro
  Contreras}.} \bibinfo{year}{2012}\natexlab{}.
\newblock \showarticletitle{Algorithms for hierarchical clustering: an
  overview}.
\newblock \bibinfo{journal}{\emph{Wiley Interdisciplinary Reviews: Data Mining
  and Knowledge Discovery}} \bibinfo{volume}{2}, \bibinfo{number}{1}
  (\bibinfo{year}{2012}), \bibinfo{pages}{86--97}.
\newblock


\bibitem[\protect\citeauthoryear{{Sun}, {Shu}, {Li}, {Yu}, {Ma}, and
  {Fang}}{{Sun} et~al\mbox{.}}{2009}]%
        {5372757}
\bibfield{author}{\bibinfo{person}{T. {Sun}}, \bibinfo{person}{C. {Shu}},
  \bibinfo{person}{F. {Li}}, \bibinfo{person}{H. {Yu}}, \bibinfo{person}{L.
  {Ma}}, {and} \bibinfo{person}{Y. {Fang}}.} \bibinfo{year}{2009}\natexlab{}.
\newblock \showarticletitle{An Efficient Hierarchical Clustering Method for
  Large Datasets with Map-Reduce}. In \bibinfo{booktitle}{\emph{2009
  International Conference on Parallel and Distributed Computing, Applications
  and Technologies}}. \bibinfo{pages}{494--499}.
\newblock
\urldef\tempurl%
\url{https://doi.org/10.1109/PDCAT.2009.46}
\showDOI{\tempurl}


\bibitem[\protect\citeauthoryear{Sun and Fox}{Sun and Fox}{2012}]%
        {sun2012study}
\bibfield{author}{\bibinfo{person}{Zhanquan Sun} {and}
  \bibinfo{person}{Geoffrey Fox}.} \bibinfo{year}{2012}\natexlab{}.
\newblock \showarticletitle{Study on parallel SVM based on MapReduce}. In
  \bibinfo{booktitle}{\emph{Proceedings of the International Conference on
  Parallel and Distributed Processing Techniques and Applications (PDPTA)}}.
  The Steering Committee of The World Congress in Computer Science,
  Computer~…, \bibinfo{pages}{1}.
\newblock


\bibitem[\protect\citeauthoryear{Tan, Steinbach, and Kumar}{Tan
  et~al\mbox{.}}{2016}]%
        {tan2016introduction}
\bibfield{author}{\bibinfo{person}{Pang-Ning Tan}, \bibinfo{person}{Michael
  Steinbach}, {and} \bibinfo{person}{Vipin Kumar}.}
  \bibinfo{year}{2016}\natexlab{}.
\newblock \bibinfo{booktitle}{\emph{Introduction to data mining}}.
\newblock \bibinfo{publisher}{Pearson Education India}.
\newblock


\bibitem[\protect\citeauthoryear{Teo, Vishwanathan, Smola, and Le}{Teo
  et~al\mbox{.}}{2010}]%
        {teo2010bundle}
\bibfield{author}{\bibinfo{person}{Choon~Hui Teo}, \bibinfo{person}{SVN
  Vishwanathan}, \bibinfo{person}{Alex Smola}, {and} \bibinfo{person}{Quoc~V
  Le}.} \bibinfo{year}{2010}\natexlab{}.
\newblock \showarticletitle{Bundle Methods for Regularized Risk Minimization.}
\newblock \bibinfo{journal}{\emph{Journal of Machine Learning Research}}
  \bibinfo{volume}{11}, \bibinfo{number}{1} (\bibinfo{year}{2010}).
\newblock


\bibitem[\protect\citeauthoryear{Veenman, Reinders, and Backer}{Veenman
  et~al\mbox{.}}{2002}]%
        {veenman2002maximum}
\bibfield{author}{\bibinfo{person}{Cor~J. Veenman}, \bibinfo{person}{Marcel
  J.~T. Reinders}, {and} \bibinfo{person}{Eric Backer}.}
  \bibinfo{year}{2002}\natexlab{}.
\newblock \showarticletitle{A maximum variance cluster algorithm}.
\newblock \bibinfo{journal}{\emph{IEEE Transactions on pattern analysis and
  machine intelligence}} \bibinfo{volume}{24}, \bibinfo{number}{9}
  (\bibinfo{year}{2002}), \bibinfo{pages}{1273--1280}.
\newblock


\bibitem[\protect\citeauthoryear{Vinh, Epps, and Bailey}{Vinh
  et~al\mbox{.}}{2009}]%
        {vinh2009information}
\bibfield{author}{\bibinfo{person}{Nguyen~Xuan Vinh}, \bibinfo{person}{Julien
  Epps}, {and} \bibinfo{person}{James Bailey}.}
  \bibinfo{year}{2009}\natexlab{}.
\newblock \showarticletitle{Information theoretic measures for clusterings
  comparison: is a correction for chance necessary?}. In
  \bibinfo{booktitle}{\emph{Proceedings of the 26th annual international
  conference on machine learning}}. \bibinfo{pages}{1073--1080}.
\newblock


\bibitem[\protect\citeauthoryear{Wagstaff, Cardie, Rogers, Schr{\"o}dl,
  et~al\mbox{.}}{Wagstaff et~al\mbox{.}}{2001}]%
        {wagstaff2001constrained}
\bibfield{author}{\bibinfo{person}{Kiri Wagstaff}, \bibinfo{person}{Claire
  Cardie}, \bibinfo{person}{Seth Rogers}, \bibinfo{person}{Stefan Schr{\"o}dl},
  {et~al\mbox{.}}} \bibinfo{year}{2001}\natexlab{}.
\newblock \showarticletitle{Constrained k-means clustering with background
  knowledge}. In \bibinfo{booktitle}{\emph{Icml}}, Vol.~\bibinfo{volume}{1}.
  \bibinfo{pages}{577--584}.
\newblock


\bibitem[\protect\citeauthoryear{Wang, Wang, and Wilkes}{Wang
  et~al\mbox{.}}{2009}]%
        {wang2009divide}
\bibfield{author}{\bibinfo{person}{Xiaochun Wang}, \bibinfo{person}{Xiali
  Wang}, {and} \bibinfo{person}{D~Mitchell Wilkes}.}
  \bibinfo{year}{2009}\natexlab{}.
\newblock \showarticletitle{A divide-and-conquer approach for minimum spanning
  tree-based clustering}.
\newblock \bibinfo{journal}{\emph{IEEE Transactions on Knowledge and Data
  Engineering}} \bibinfo{volume}{21}, \bibinfo{number}{7}
  (\bibinfo{year}{2009}), \bibinfo{pages}{945--958}.
\newblock


\bibitem[\protect\citeauthoryear{Wikipedia}{Wikipedia}{2020}]%
        {c.elmohamed}
\bibfield{author}{\bibinfo{person}{Wikipedia}.}
  \bibinfo{year}{2020}\natexlab{}.
\newblock \bibinfo{title}{Global city}.
\newblock \bibinfo{howpublished}{Website}.
\newblock
\newblock
\shownote{\url{https://en.wikipedia.org/wiki/Global_city\#Global_Cities_Index}.}


\bibitem[\protect\citeauthoryear{Yang, Luan, Li, and Qian}{Yang
  et~al\mbox{.}}{2012}]%
        {yang2012mapreduce}
\bibfield{author}{\bibinfo{person}{Hailong Yang}, \bibinfo{person}{Zhongzhi
  Luan}, \bibinfo{person}{Wenjun Li}, {and} \bibinfo{person}{Depei Qian}.}
  \bibinfo{year}{2012}\natexlab{}.
\newblock \showarticletitle{MapReduce workload modeling with statistical
  approach}.
\newblock \bibinfo{journal}{\emph{Journal of grid computing}}
  \bibinfo{volume}{10}, \bibinfo{number}{2} (\bibinfo{year}{2012}),
  \bibinfo{pages}{279--310}.
\newblock


\bibitem[\protect\citeauthoryear{You}{You}{2020}]%
        {you2020fast}
\bibfield{author}{\bibinfo{person}{Yang You}.} \bibinfo{year}{2020}\natexlab{}.
\newblock \showarticletitle{Fast and Accurate Machine Learning on Distributed
  Systems and Supercomputers}.
\newblock  (\bibinfo{year}{2020}).
\newblock


\bibitem[\protect\citeauthoryear{You, Demmel, Hsieh, and Vuduc}{You
  et~al\mbox{.}}{2018}]%
        {you2018accurate}
\bibfield{author}{\bibinfo{person}{Yang You}, \bibinfo{person}{James Demmel},
  \bibinfo{person}{Cho-Jui Hsieh}, {and} \bibinfo{person}{Richard Vuduc}.}
  \bibinfo{year}{2018}\natexlab{}.
\newblock \showarticletitle{Accurate, fast and scalable kernel ridge regression
  on parallel and distributed systems}. In
  \bibinfo{booktitle}{\emph{Proceedings of the 2018 International Conference on
  Supercomputing}}. \bibinfo{pages}{307--317}.
\newblock


\bibitem[\protect\citeauthoryear{Zahn}{Zahn}{1971}]%
        {zahn1971graph}
\bibfield{author}{\bibinfo{person}{Charles~T Zahn}.}
  \bibinfo{year}{1971}\natexlab{}.
\newblock \showarticletitle{Graph-theoretical methods for detecting and
  describing gestalt clusters}.
\newblock \bibinfo{journal}{\emph{IEEE Transactions on computers}}
  \bibinfo{volume}{100}, \bibinfo{number}{1} (\bibinfo{year}{1971}),
  \bibinfo{pages}{68--86}.
\newblock


\bibitem[\protect\citeauthoryear{Zhang, Duchi, and Wainwright}{Zhang
  et~al\mbox{.}}{2013}]%
        {zhang2013divide}
\bibfield{author}{\bibinfo{person}{Yuchen Zhang}, \bibinfo{person}{John Duchi},
  {and} \bibinfo{person}{Martin Wainwright}.} \bibinfo{year}{2013}\natexlab{}.
\newblock \showarticletitle{Divide and conquer kernel ridge regression}. In
  \bibinfo{booktitle}{\emph{Conference on learning theory}}.
  \bibinfo{pages}{592--617}.
\newblock


\bibitem[\protect\citeauthoryear{Zhang, Duchi, and Wainwright}{Zhang
  et~al\mbox{.}}{2015}]%
        {zhang2015divide}
\bibfield{author}{\bibinfo{person}{Yuchen Zhang}, \bibinfo{person}{John Duchi},
  {and} \bibinfo{person}{Martin Wainwright}.} \bibinfo{year}{2015}\natexlab{}.
\newblock \showarticletitle{Divide and conquer kernel ridge regression: A
  distributed algorithm with minimax optimal rates}.
\newblock \bibinfo{journal}{\emph{The Journal of Machine Learning Research}}
  \bibinfo{volume}{16}, \bibinfo{number}{1} (\bibinfo{year}{2015}),
  \bibinfo{pages}{3299--3340}.
\newblock


\bibitem[\protect\citeauthoryear{Zhong, Malinen, Miao, and Fr{\"a}nti}{Zhong
  et~al\mbox{.}}{2015}]%
        {zhong2015fast}
\bibfield{author}{\bibinfo{person}{Caiming Zhong}, \bibinfo{person}{Mikko
  Malinen}, \bibinfo{person}{Duoqian Miao}, {and} \bibinfo{person}{Pasi
  Fr{\"a}nti}.} \bibinfo{year}{2015}\natexlab{}.
\newblock \showarticletitle{A fast minimum spanning tree algorithm based on
  K-means}.
\newblock \bibinfo{journal}{\emph{Information Sciences}}  \bibinfo{volume}{295}
  (\bibinfo{year}{2015}), \bibinfo{pages}{1--17}.
\newblock


\end{thebibliography}



\end{document}